\documentclass[conference]{IEEEtran}
\IEEEoverridecommandlockouts
\usepackage{cite}
\usepackage{amsmath,amssymb,amsfonts}
\usepackage{graphicx}
\usepackage{textcomp}
\usepackage{xcolor}

\usepackage{multirow}
\newcommand{\tabincell}[2]{\begin{tabular}{@{}#1@{}}#2\end{tabular}}  

\usepackage{float}
\usepackage{subcaption}

\usepackage{algpseudocode}
\usepackage[linesnumbered, ruled, vlined]{algorithm2e}
\usepackage{amsmath,amsthm,amssymb}
\renewcommand{\algorithmicrequire}{ \textbf{Input: }} 
\renewcommand{\algorithmicensure}{ \textbf{Output: }} 

\usepackage{url}

\usepackage{pifont}
\newcommand{\cmark}{\ding{51}}
\newcommand{\xmark}{\ding{55}}

\newcommand{\Name}{{Fast-BNS}}

\def\BibTeX{{\rm B\kern-.05em{\sc i\kern-.025em b}\kern-.08em
    T\kern-.1667em\lower.7ex\hbox{E}\kern-.125emX}}
\begin{document}

\title{Fast Parallel Bayesian Network Structure Learning}


\author{\IEEEauthorblockN{Jiantong Jiang,
Zeyi Wen\IEEEauthorrefmark{1}\thanks{\IEEEauthorrefmark{1} Zeyi Wen is the corresponding author.} and
Ajmal Mian}
\IEEEauthorblockA{\textit{Department of Computer Science and Software Engineering, The University of Western Australia} \\ jiantong.jiang@research.uwa.edu.au, zeyi.wen@uwa.edu.au, ajmal.mian@uwa.edu.au}}

\maketitle

\begin{abstract}
Bayesian networks (BNs) are a widely used graphical model in machine learning for representing knowledge with uncertainty. The mainstream BN structure learning methods require performing a large number of conditional independence (CI) tests. The learning process is very time-consuming, especially for high-dimensional problems, which hinders the adoption of BNs to more applications. 
Existing works attempt to accelerate the learning process with parallelism, but face issues including load unbalancing, costly atomic operations and dominant parallel overhead. In this paper, we propose a fast solution named \Name{} on multi-core CPUs to enhance the efficiency of the BN structure learning. \Name{} is powered by a series of efficiency optimizations including (i) designing a dynamic work pool to monitor the processing of edges and to better schedule the workloads among threads, (ii) grouping the CI tests of the edges with the same endpoints to reduce the number of unnecessary CI tests, (iii) using a cache-friendly data storage to improve the memory efficiency, and (iv) generating the conditioning sets on-the-fly to avoid extra memory consumption. A comprehensive experimental study shows that the sequential version of \Name{} is up to 50 times faster than its counterpart, and the parallel version of \Name{} achieves 4.8 to 24.5 times speedup over the state-of-the-art multi-threaded solution. Moreover, \Name{} has a good scalability to the network size as well as sample size.
\Name{} source code is freely available at \url{https://github.com/jjiantong/FastBN}.

\end{abstract}


\section{Introduction}
Bayesian networks (BNs)~\cite{pearl1988bn} are probabilistic graphical models that employ directed acyclic graphs (DAGs) to compactly represent a set of random variables and their conditional dependency. 
The graphical nature of BNs makes them well-suited for representing knowledge with uncertainty and efficient reasoning. They have been successfully applied in a wide range of real-world applications~\cite{kyrimi2021comprehensive, ramsey2017million, marcot2019advances, sun2018using}.
With recent growing demand for interpretable machine learning models, BNs have attracted much research attention since they are inherently interpretable models~\cite{gunning2019darpa, rudin2019stop}. 

One crucial task of training BNs is structure learning, which aims to learn DAGs that are well matched the observed data. There are two common approaches for BN structure learning from data: score-based approaches and constraint-based approaches. 
The score-based approaches use a scoring function to measure the fitness of DAGs to the data and find the highest score out of all the possible DAGs, which makes the number of possible DAGs super-exponential to the number of dimensions (i.e., variables) of the learning problems~\cite{robinson1977counting}. 
On the other hand, the constraint-based approaches perform a number of conditional independence (CI) tests to identify the conditional independence relations among the random variables and use these relations as constraints to construct BNs. This category of methods often runs in a polynomial time, and is commonly used in high-dimensional problems~\cite{kalisch2007estimating}.


A fundamental constraint-based algorithm is the PC (named after its authors Peter and Clark) algorithm~\cite{spirtes2000causation} which starts from a complete undirected graph and removes edges in consecutive depths based on CI tests.
PC-stable~\cite{colombo2014order} solves the order-dependent issue in the original PC algorithm and produces less error. The PC-stable algorithm has been widely used in various applications~\cite{zhang2012inferring, maathuis2010predicting} and is implemented in different mainstream BN packages such as bnlearn~\cite{scutari2009learning}, pcalg~\cite{kalisch2012causal} and tetrad~\cite{ramsey2018tetrad}.
Furthermore, most constraint-based methods are improved versions of the PC-stable algorithm or proceed along similar lines of the PC-stable algorithm.

However, a key barrier that hinders the wider usage of the PC-stable algorithm is its long execution time for performing a large number of CI tests, especially for high-dimensional data sets. It is non-trivial to perform algorithmic improvements for the PC-stable algorithm~\cite{scutari2019learns}.
Several research works have been conducted on the acceleration of the PC-stable algorithm on multi-core CPUs exploiting parallelization techniques~\cite{scutari2014bayesian, madsen2017parallel, le2016fast}. The most common way is to parallelize the processing of different edges of the network inside each depth, which is an intuitive idea due to the order-independent property of the PC-stable algorithm. However, the direct edge-level parallelism is load unbalanced, because the workloads of CI tests for different edges is highly different. Another approach to parallelize the algorithm is by processing multiple samples inside each CI test, which is a finer granularity of parallelism. However, this approach requires many atomic operations and has a large parallel overhead, which decreases the efficiency.

To address the issues of workload balancing, atomic operations and large parallel overhead, we propose \Name{}, a fast BN structure learning solution. \Name{} is equipped with a dynamic work pool to contain the edges to be processed and their processing progresses with regard to the CI tests. 
The work pool is able to monitor the processing progresses of edges, terminating the completed edges in time to avoid unnecessary CI tests.
Moreover, with the work pool, we can better schedule the work among threads to maintain workload balancing. Each thread always processes a group of CI tests that are required to be processed, and hence all the threads are active during this process. As the CI tests are independent and each CI test requires a reasonable amount of computation, \Name{} can be performed without atomic operations and can amortize the parallel overhead. To further enhance \Name{}, we develop a series of efficiency optimizations including (i) grouping the CI tests of the edges with the same endpoints to reduce the number of CI tests, (ii) using a cache-friendly data storage to improve memory efficiency and (iii) generating the conditioning sets of the CI tests on-the-fly and in parallel to avoid extra memory consumption. To summarize, we make the following major contributions in this paper.

\begin{itemize}
\item We propose \Name{} to accelerate the BN structure learning on multi-core CPUs. \Name{} employs a dynamic work pool to monitor the processing of edges and schedule the workloads among threads, which solves the efficiency issues of load unbalancing. The granularity of parallelism used in \Name{} avoids atomic operations and amortizes the parallel overhead.
\item We develop a series of novel techniques to further improve the efficiency of \Name{}. First, we propose to group the CI tests of the edges with the same endpoints together to reduce the number of unnecessary CI tests. Second, we employ a cache-friendly data storage to improve the memory efficiency. Lastly, we compute the conditioning sets on-the-fly and in parallel to reduce memory consumption.
\item We conduct experiments to study the effectiveness of our proposed techniques. Experimental results show that the sequential version of \Name{} outperforms the existing work bnlearn~\cite{scutari2009learning} and tetrad~\cite{ramsey2018tetrad} by up to 50 times. When compared with the multi-threaded implementation in bnlearn~\cite{scutari2014bayesian}, the parallel version of \Name{} is 4.8 to 24.5 times faster. 
Finally, we show that \Name{} has good scalability to the network size and sample size.
\end{itemize}


\section{Related Work}
\label{sec_related}

Bayesian Networks (BNs) are powerful models for representation learning and reasoning under uncertainty in artificial intelligence. BNs have recently attracted much attention within the research and industry communities. A crucial aspect of using BNs is to learn the dependency graph of a BN from data, which is called \emph{structure learning}. In this paper, we categorize the related work on BN structure learning into 
two groups: score-based approaches and constraint-based approaches.

\emph{Score-based approaches}~\cite{chickering2002optimal, chickering1995learning, acid2003searching, larranaga1996structure, tian2013branch, myers2013learning, blanco2003learning} seek the best directed acyclic graph (DAG) according to scoring functions that measure the fitness of BN structures to the observed data. Widely adopted scores include BDeu, BIC, and MDL
.
However, the number of possible DAGs is super-exponential to the number of variables~\cite{robinson1977counting}.
Hence, many score-based approaches employ heuristic methods, like greedy search or simulated annealing, in an attempt to reduce the search space. Such approaches can easily get trapped in local optima \cite{scutari2019learns}. The optimization techniques in this paper focus on the constraint-based approaches which tend to scale better to high-dimensional data.

\emph{Constraint-based approaches}~\cite{spirtes2000causation, colombo2014order, colombo2012learning, richardson2013discovery, harris2013pc, margaritis2003learning, yaramakala2005speculative} perform structure learning using a series of statistical tests, such as Chi-square test, $G^2$ test and mutual information test, to learn the conditional independence relationships among the variables in the model. The DAG is then built according to these relations as constraints. Most of the constraint-based algorithms proceed along similar lines as the work of the PC-stable 
algorithm~\cite{spirtes2000causation, colombo2014order}. 
Unlike score-based approaches, it is generally non-trivial to perform algorithmic improvements for constraint-based approaches using general-purpose optimization theory. This paper mainly focuses on improving the efficiency of the PC-stable algorithm using parallel techniques.


There are some well-known open-source BN libraries which contain the implementation of the PC-stable algorithm, such as bnlearn~\cite{scutari2009learning}, pcalg~\cite{kalisch2012causal} and tetrad~\cite{ramsey2018tetrad}. 
Meanwhile, since the recent parallel computing platforms, such as multi-core CPUs and GPUs, have emerged to efficiently address various computational machine learning problems~\cite{wen2018thundersvm, Jiang2022ParallelAD}, there are several research works that focus on the acceleration of the PC-stable algorithm using parallel techniques~\cite{scutari2014bayesian, madsen2017parallel, le2016fast}. The key idea is to parallelize the processing of different edges inside each depth, which is an intuitive idea due to the order-independent property of the PC-stable algorithm. However, the edge-level parallelism is load unbalanced, because the workload of the conditional independence tests for different edges is highly skewed. This paper improves the efficiency of the PC-stable algorithm using CI-level parallelism to boost the applications of the BN structure learning.

\section{Preliminaries}
\label{sec_pre}
In this section, we provide the key terminologies and definitions related to Bayesian Network structure learning, and then review the PC-stable algorithm.

\subsection{Bayesian Networks}

Bayesian Networks (BNs) are a class of graphical models that represent a joint distribution $P$ over a set of random variables $\mathcal{V} = \{V_0, V_1, ... , V_{n-1}\}$ via a directed acyclic graph (DAG). Typically, one variable corresponds to one feature in the machine learning problems.
We use $G = (\mathcal{V}, \mathcal{E})$ to denote the DAG. In a DAG $G$, each node in $\mathcal{V}$ is associated with one variable and each edge in $\mathcal{E}$ represents conditional dependencies among the two variables. $V_j$ is called a parent of $V_i$ if there exists a directed edge from $V_j$ to $V_i$ in $G$, and we use $Pa(V_i)$ to denote the set of parent variables of $V_i$.

In a BN, each variable has its local probability distribution
that describes the probabilities of possible values of this variable given its possible parent configurations. The joint probability of variables $\mathcal{V}$ in a BN can be decomposed into the product of the local probability distributions of each variable, and each local probability distribution depends only on a single variable $V_i$ and its parents:
\begin{equation*} 
\label{equ_chain_rule}
P(V_0, V_1, ...,V_{n-1}) = \prod_{i=0}^{n-1} P(V_i | Pa(V _i))
\end{equation*}
where $n$ is the number of variables, $P(V_0, V_1, ...,V_{n-1})$ is the joint probability and $P(V_i | Pa(V_i))$ is the conditional probability of variable $V_i$.

\subsection{Conditional Independence Tests}

Consider some random variables $V_i$, $V_j$ and $V_k$ in a BN, a CI test assertion of the form $I(V_i, V_j | \{V_k\})$ means $V_i$ and $V_j$ are independent given $V_k$. Let $\mathcal{D} = \{c_0, c_1, ..., c_{m-1}\}$ denote a data set of $m$ complete samples, a CI test $I(V_i, V_j | \{V_k\})$ determines whether the corresponding hypothesis $I(V_i, V_j | \{V_k\})$ holds or not, based on statistics of $D$. For discrete variables, the most common statistic for testing $I(V_i, V_j | \{V_k\})$ is the $G^2$ test statistic~\cite{spirtes2000causation} defined as
\begin{equation*} 
\label{equ_g2}
G^2 = 2 \sum_{x, y, z} N_{xyz} log \frac{N_{xyz}}{E_{xyz}},
\end{equation*}
where $N_{xyz}$ is the number of samples in $D$ that satisfy $V_i = x$, $V_j = y$ and $V_k = z$. The value of $N_{xyz}$ can be obtained from the contingency table that shows the frequencies for all configurations of values. 
$G^2$ follows an asymptotic $\chi^2$ distribution with $(|V_i|-1)(|V_j|-1)$,
where $|\cdot|$ denotes the number of possible values of the variable. The p value of $\chi^2$ distribution can be calculated according to the $G^2$ statistic and the final decision is made by comparing p value with the significance level $\alpha$. If p value is greater than $\alpha$, the independent hypothesis $I(V_i, V_j | \{V_k\})$ is accepted; otherwise, the hypothesis is rejected.
$E_{xyz}$ is the expected frequency which is defined as
\begin{equation*} 
\label{equ_e}
E_{xyz} = \frac{N_{x+z} N_{+yz}}{N_{++z}},
\end{equation*}
where $N_{x+z} = \sum_{y} N_{xyz}$, $N_{+yz} = \sum_{x} N_{xyz}$, and $N_{++z} = \sum_{xy} N_{xyz}$, which represent the marginal frequencies.

\subsection{The PC-Stable Algorithm}

The PC-stable 
algorithm is a constraint-based method for BN structure learning from data. PC-stable consists of three steps. The first step is to determine the skeleton of the graph. The term skeleton means the underlying undirected graph of the learned network. This step is done by performing a large number of CI tests. The second step is to identify the v-structures in the skeleton. A v-structure is a triple $(V_i, V_j, V_k)$ that can be denoted by $V_i \rightarrow V_k \leftarrow V_j$. In other words, nodes $V_i$ and $V_j$ have an outgoing edge to node $V_k$ and are not connected by any edge in the graph. 
V-structure is a key component to distinguish different network structures. By identifying the v-structures in this step, some edges in the skeleton become directed edges. The third step is to set directions for as many of the remaining undirected edges as possible by applying a set of rules called Meek rules~\cite{meek2013causal}. For example, we set the direction of the undirected edge $V_j - V_k$ into $V_j \rightarrow V_k$ whenever there is a directed edge $V_i \rightarrow V_j$ such that $V_i$ and $V_k$ are not adjacent; otherwise a new v-structure is created.
In the three steps of the PC-stable algorithm, the first step is much more time-consuming~\cite{zarebavani2019cupc}, taking more than 90\% of the total execution time in many problems. In Section~\ref{paper:alg}, we elaborate the details of our proposed techniques to accelerating the first step. 

The pseudo-code of the first step of the PC-stable algorithm is given in Algorithm~\ref{alg_pc_stable}. 
The general idea is to initialize the graph $G$ to a complete undirected graph over the node set $\mathcal{V}$ (Line 3), and remove some of the edges by performing a number of CI tests in consecutive depths (Lines 5 to 20).
Specifically, at each depth $d$, the algorithm iteratively records the current adjacency sets of all the nodes (Lines 6 to 8), where $adj(G, V_i)$ denotes the adjacent nodes of $V_i$ in $G$. This operation is used for choosing the conditioning set $\mathcal{S}$ later.
Next, for every edge $(V_i, V_j)$ in the graph $G$, a number of CI tests $I(V_i, V_j | \mathcal{S})$ are performed for different conditioning sets. The elements in the conditioning sets are chosen from $a(V_i)\backslash\{V_j\}$, and the size of each conditioning set $|\mathcal{S}|$ is equal to the current depth $d$ (Lines 9 to 12). If there exists a conditioning set $\mathcal{S}$ where $V_i$ is independent of $V_j$ given $\mathcal{S}$, the edge $(V_i, V_j)$ is removed from $G$, and $\mathcal{S}$ is stored in $SepSet(V_i, V_j)$ (Lines 13 to 15). $SepSet(V_i, V_j)$ denotes the separating set of $V_i$ and $V_j$, which is used in the second step of the PC-stable algorithm to identify the v-structures. Since the second step is fast and is not the focus of our work, we omit the details of separating set. Once all edges are considered, $d$ is incremented (Line 19) and the above procedure is repeated for the next depth. Depth $d$ is used to control the size of the conditioning sets from small to large. 
This process continues until all pairs of adjacent nodes $(V_i, V_j)$ in $G$ satisfy $|a(V_i)\backslash\{V_j\}| < d$ as shown in Line 20.

\begin{algorithm} [tb]
\DontPrintSemicolon
\LinesNumbered
\caption{The first step of the PC-stable algorithm.}   
\label{alg_pc_stable}

\algorithmicrequire Node set $\mathcal{V}$

\algorithmicensure  Graph $G$, $SepSet$

Form the complete undirected graph $G$ over $\mathcal{V}$

Let depth $d = 0$
 
\textbf{repeat} 

\quad \textbf{for} any node $V_i$ in $G$ \textbf{do} 
 
\quad \quad Let $a(V_i) = adj(G, V_i)$

\quad \textbf{end for}

\quad \textbf{for} any edge $(V_i, V_j)$ in $G$ \textbf{do} 
 
\quad \quad \textbf{repeat}

\quad \quad \quad Choose a new $\mathcal{S} \subseteq a(V_i)\backslash\{V_j\}$ with $|\mathcal{S}| = d$ 

\quad \quad \quad Perform CI test $I(V_i, V_j | \mathcal{S})$

\quad \quad \quad \textbf{if} hypothesis $I(V_i, V_j | \mathcal{S})$ holds \textbf{then}

\quad \quad \quad \quad Remove $(V_i, V_j)$ from $G$

\quad \quad \quad \quad Store $\mathcal{S}$ in $SepSet(V_i, V_j)$

\quad \quad \quad \textbf{end if}

\quad \quad \textbf{until} $(V_i, V_j)$ is removed or all $\mathcal{S}$ are considered

\quad \textbf{end for}

\quad Let $d = d + 1$

\textbf{until} all pairs of $(V_i, V_j)$ in $G$ satisfy $|a(V_i)\backslash\{V_j\}| < d$
 
\end{algorithm}

\section{Parallel BN Structure Learning}
\label{paper:alg}

This section elaborates the technical details of our proposed \Name{}.
First, we identify two granularities of parallelism, including edge-level parallelism and sample-level parallelism.
We find some shortcomings of accelerating the PC-stable algorithm using the above two granularities of parallelism, including (i) load unbalancing between threads, (ii) many atomic operations and (iii) high parallel overhead. To remedy the shortcomings, we propose \Name{} which exploits a CI-level parallelism, where multiple groups of CI tests from different edges are performed in parallel. \Name{} takes advantage of a dynamic work pool that contains the edges required to be processed and their processing progress to ensure load balancing among threads. The CI-level parallelism also avoids atomic operations and leads to a reasonable amount of workloads to amortize the overhead of parallel computing. To further enhance the efficiency of \Name{}, we propose a series of novel optimizations including (i) grouping the CI tests to reduce unnecessary CI tests, (ii) employing a cache-friendly data storage to improve the memory efficiency, and (iii) generating the conditioning sets of the CI tests on-the-fly and in parallel to avoid extra memory consumption.

\subsection{Edge-Level and Sample-Level Parallelism}
\label{sec_para}

Here, we first describe two schemes using two granularities of parallelism: edge-level parallelism and sample-level parallelism. Then, we discuss the limitations of these schemes.

\begin{figure}
	\centering
	\includegraphics[width=0.48\textwidth]{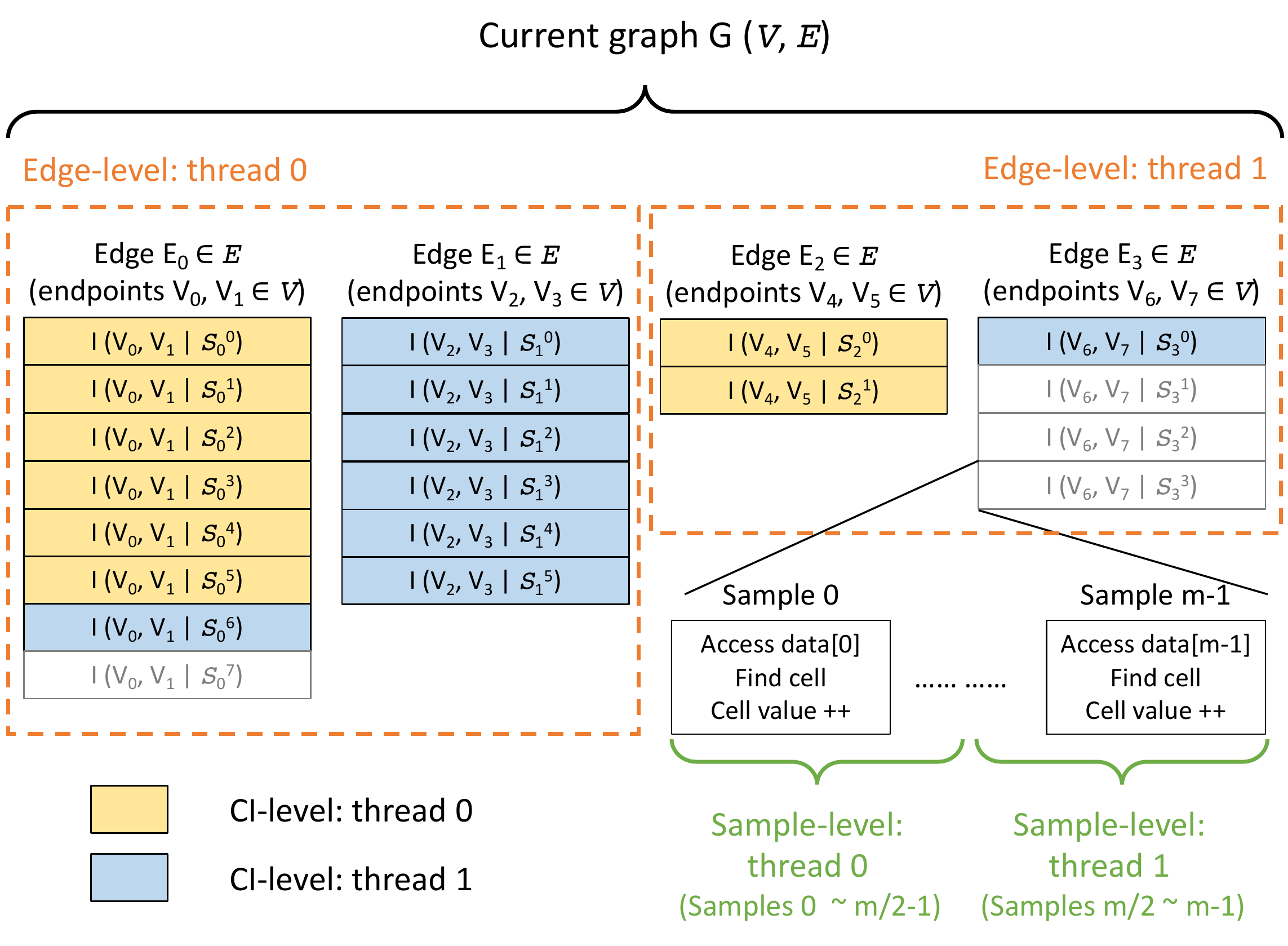}
	\caption{Three 
	different granularities of parallelism: edge-level parallelism, sample-level parallelism and CI-level parallelism.}
	\label{fig_overview}
\end{figure}

\emph{Edge-Level Parallelism:} 
The most natural scheme to parallelize the PC-stable algorithm is to parallelize the processing of different edges inside each depth, which is a coarse-grained parallelism. In each depth $d$, it parallelizes the for-loop in Line 9 of Algorithm~\ref{alg_pc_stable}, dedicating $\frac{|\mathcal{E}_d|}{t}$ edges to each thread, where $t$ represents the number of threads and $|\mathcal{E}_d|$ represents the number of edges to be processed in the depth $d$. A simple example is shown in Figure~\ref{fig_overview}. The current graph contains four edges, and thus for the case of $t = 2$, each of the two threads is responsible for processing two edges. Specifically, thread 0 is dedicated to edges $E_0$ and $E_1$, while $E_2$ and $E_3$ are assigned to thread 1.
This is an intuitive idea because the order-independent property of PC-stable makes it suitable for parallelizing at each depth. In other words, an edge deletion does not effect the processing of other edges at the same depth, and thus parallelization can be applied.

\emph{Sample-Level Parallelism.} 
Another scheme is to parallelize among samples inside each CI test, which is a fine-grained parallelism. Since there are a large number of CI tests in the PC-stable algorithm and they take most of the execution time, the key reason behind this scheme is to optimize each of the CI tests.
To explain the sample-level parallelism, we decompose the processing of CI tests into three steps: (i) generating the contingency table, (ii) computing the marginal tables, and (iii) computing $G^2$ statistics and determining the dependency hypothesis. 
Among them, the most important step is to generate the contingency table.
In this step, the contingency table is generated by traversing the whole data set. Specifically, for the CI test $I(V_i, V_j | \{V_k\})$, it accesses all samples in the data set and finds one cell in the contingency table for each sample by getting the values of $V_i$, $V_j$ and $V_k$ of the sample, and finally increments one to this cell.
Therefore, the sample-level parallelism is to parallelize the traversing of the whole data set, dedicating $\frac{m}{t}$ samples to each thread, where $m$ represents the number of samples in the data set. For the example in Figure~\ref{fig_overview}, for the case of $t = 2$, each thread would process $\frac{m}{2}$ samples for every CI test.

\emph{Limitations of Edge-Level Parallelism. }
The coarse-grained edge-level parallelism is load unbalanced, because the workloads of CI tests for different edges is highly different due to the following reasons. 
\begin{itemize}
    \item Firstly, different nodes in the network may have different number of adjacent nodes. For example, if node $V_i$ has two adjacent nodes while $V_k$ has ten adjacent nodes, then there is $\binom{2}{2} = 1$ conditioning set when processing edge $V_i - V_j$ in the depth $d=2$, while there are $\binom{10}{2} = 45$ possible conditioning sets for edge $V_k - V_j$ in the same depth. We can see from this example that the number of possible conditioning sets can be quite different. As shown in Figure~\ref{fig_overview}, edges $E_0$ and $E_1$ assigned to thread 0 have 14 possible CI tests in total, while $E_2$ and $E_3$ assigned to thread 1 have only 6 possible CI tests.
    \item Secondly, when a hypothesis $I(V_i, V_j | \mathcal{S})$ holds, the edge between $V_i$ and $V_j$ is removed in advance. Then the CI tests conditioning on the remaining conditioning sets are unnecessary. Let us take the edge $E_3$ in Figure~\ref{fig_overview} as an example. If $V_6$ and $V_7$ are conditionally independent given the first conditioning set tested (i.e., $\mathcal{S}_3^0$), the edge $E_3$ is removed in advance and the three CI tests of $V_6$ and $V_7$ left are unnecessary to perform, as there is no edge between $V_6$ and $V_7$ now. Such unnecessary CI tests are marked in gray in Figure~\ref{fig_overview}.
    This example shows that we do not know in advance as to how many CI tests are required for each edge, since any edge has the possibility of being removed before finishing its CI tests conditioning on all its possible conditioning sets.
\end{itemize}

\emph{Limitations of Sample-Level Parallelism:}
The sample-level parallelism is fine-grained, but fails to get good speedups due to the following two main reasons.
\begin{itemize}
    \item Firstly, since each sample contributes to the generation of the contingency table, sample-level parallelism may result in a race condition when updating the cells in the contingency table. Therefore, we need many atomic operations to guarantee the correctness of the execution. However, atomic operations are expensive. An alternative way to perform the contingency table generation is to create a local contingency table for each thread. For a specific CI test, each thread would then be responsible for updating its local table according to its own data. Then the local tables from all threads are combined to generate the final contingency table. However, this alternative approach requires much more memory and introduces non-negligible costs of frequent synchronization and communication among threads.
    \item Secondly, although generating the contingency table is the most time-consuming step, it is mainly due to the large number of such operations, while the workload of each operation is relatively small. We need a larger amount of workload for each thread in order to amortize the overhead of parallel computing (e.g., thread creation).
\end{itemize}

\subsection{CI-Level Parallelism}
\label{sec_ours}

To overcome the limitations of the edge-level and sample-level parallelism, our proposed \Name{} employs a CI-level parallelism, which has a parallelism granularity between edge-level and sample-level. 
In each depth, CI tests from different edges are performed in parallel. To achieve CI-level parallelism, our key idea is to employ a dynamic work pool for each depth implemented by a stack. The work pool contains the edges required to be processed and their processing progresses with respect to the CI tests. Therefore, each time we can fetch multiple edges required to be processed from the work pool, find the next groups of CI tests of the edges through their processing progresses, and execute them in parallel.

In particular, at the beginning of each depth, all the edges in the current graph $G$ are pushed into the work pool with zero processing progress. Then, each time $t$ edges are popped from the work pool and assigned to the threads. Each thread would then be responsible for processing a group of CI tests of the assigned edge, where the number of CI tests is introduced as $gs$ ($gs \geq 1$). When the $gs$ CI tests are finished, 
Two decisions are made, including whether to accept the independence hypothesis of the group of CI tests and whether the edge is required to be pushed into the work pool.
Specifically, the independence hypothesis of the group is accepted if any one of the CI tests in the group accepts its independence hypothesis; otherwise, the hypothesis is rejected. If the independence hypothesis of the group is accepted, or the edge has finished all its CI tests after processing this group, it means that the processing of the edge is completed, and thus the edge does not need to be pushed back to the work pool; otherwise, the edge would be pushed back to the work pool with its processing progress recorded as the last processed CI test. After that, $t$ edges are popped from the work pool, and the next $gs$ CI tests (according to the processing progress) of each edge are processed by $t$ parallel threads again. This process is performed iteratively until the work pool is empty.

Intuitively, one can think of this process as multiple threads processing multiple CI tests on different edges in parallel, but a thread is not bounded to a fixed edge. 
When the processing of an edge is finished, the thread turns to process the CI tests of other edges immediately without waiting for other edges to be finished.
This is due to the design of dynamic work pool that monitors the processing progress of each edge. With the edge monitoring technique, the completed edges are terminated in time to reduce the unnecessary CI tests.
Moreover, we can better schedule the workloads among threads with the help of the design of the dynamic work pool.
All the threads always process the CI tests that are required to be processed, and hence all the threads are active in the parallel region. 
As shown in Figure~\ref{fig_overview}, the CI tests in yellow are scheduled to thread 0 and the ones in blue are scheduled to thread 1. The CI tests of one edge are not necessarily processed by one thread.

Compared with the edge-level parallelism, the proposed CI-level parallelism has a finer granularity. 
The workloads of CI tests can be evenly distributed to each thread with the design of the dynamic work pool, which solves the issue of load unbalancing.
On the other hand, CI-level parallelism is coarser than the sample-level parallelism. Since one CI test is not distributed to multiple threads, each thread holds a complete contingency table and the race condition does not occur. Therefore, the CI-level parallelism avoids expensive atomic operations and also has a reasonable amount of workloads to amortize the parallel overhead.
Table~\ref{tab_para} summarizes the key differences between edge-level parallelism, sample-level parallelism and the CI-level parallelism.

\begin{table}
    \centering
    \caption{Comparison between edge-level parallelism, sample-level parallelism and the proposed CI-level parallelism.}
    \begin{tabular}{|c|c|c|c|}
        \hline
        Granularity of parallelism & \tabincell{c}{Load\\balance} & \tabincell{c}{Avoid atomic\\operations} & \tabincell{c}{Reasonable\\workloads} \\
        \hline
        Edge-level parallelism & \xmark & \cmark & \cmark \\
        Sample-level parallelism & \cmark & \xmark & \xmark \\
        CI-level parallelism & \cmark & \cmark & \cmark \\
        \hline
    \end{tabular}
    \label{tab_para}
\end{table}

The $gs$ in the CI-level parallelism is a trade-off between the number of CI tests and memory accesses. 
In the parallel region, each thread processes $gs$ CI tests of the same edge $V_i - V_j$ each time and makes the decision according to the results. Hence, the CI tests in a group share the same form of $I(V_i, V_j | \mathcal{S}_n)$, $0 \leq n < gs$. Since $V_i$ and $V_j$ are common for the whole group, we propose to reuse them to reduce the memory accesses when traversing the data set. The reduced memory accesses increase as the increase of $gs$. 
However, more redundant CI tests are introduced at the same time, because all the CI tests in a group are required to be performed before making the final decision on whether the edge is required to be processed again. In a special case of $gs = 1$, no redundant CI tests are introduced.
We carefully examine the effect of $gs$ and observe that some small $gs$ like 6 or 8 are good choices in practice.

It is worth noting that the CI-level parallelism is used when the depth $d \geq 1$. In depth $d = 0$, the conditioning set $S = \emptyset$ as the size of conditioning sets is equal to $d$ (cf. Algorithm~\ref{alg_pc_stable}, Line 11). Specifically, for each edge $(V_i, V_j)$ in $G$, only one CI test is required, which is $I(V_i, V_j | \emptyset)$ or simply a marginal independence test $I(V_i, V_j)$. In other words, we know in advance how many CI tests are required in depth zero, which is equal to $n(n-1)/2$, representing the number of edges in the complete undirected graph $G$ over the node set $\mathcal{V}$, where $n$ represents the number of nodes. Consequently, 
the required computations for depth zero can be simplified. Therefore, the direct edge-level parallelism is applied to depth zero without the efficiency issue of load unbalancing.

\subsection{Further Enhancing \Name{}}
\label{sec_opts}

We find three issues in existing implementations. 
First, it is inefficient to distribute the CI tests of the edges with the same endpoints to different threads, as it may cause unnecessary CI tests.
Second, the memory access pattern is irregular because the required values of one CI test are not necessarily stored sequentially.
Third, the number of CI tests is large, which requires much memory to store the indices of the conditioning sets for all the CI tests.
In this section, we aim to tackle these three issues to further improve the overall efficiency. 


\emph{Grouping CI tests of the edges with the same endpoints:}
We view the edges with the same endpoints, such as edges $V_i - V_j$ and $V_j - V_i$, as the same edge in \Name{}, instead of separating them as in the original PC-stable algorithm, because it is inefficient to separate the CI tests of two such edges. For instance, given the edge between $V_i$ and $V_j$, we need to perform the CI tests conditioning on the variables in $adj(G, V_i) \backslash \{V_j\}$ and $adj(G, V_j) \backslash \{V_i\}$. However, if we first perform the CI tests conditioning on the variables in $adj(G, V_i) \backslash \{V_j\}$ and the edge between $V_i$ and $V_j$ is removed, then the CI tests conditioning on variables in $adj(G, V_j) \backslash \{V_i\}$ are unnecessary. Therefore, we solve this dependency by grouping the CI tests of the edges with the same endpoints together to reduce the number of CI tests to be performed, and thus improve the efficiency. If the CI tests between $V_i$ and $V_j$ accept the independence hypothesis when conditioning on $S \in adj(G, V_i) \backslash \{V_j\}$, \Name{} does not perform the CI tests conditioning on the variables in $adj(G, V_j) \backslash \{V_i\}$.

\emph{Using a cache-friendly data storage:}
As discussed in Section~\ref{sec_para}, a key step of the algorithm is to compute the contingency table. For example, to test $I(X, Y| \{Z_1, Z_2\})$, we need to traverse the whole data set and obtain the values of $X$, $Y$, $Z_1$ and $Z_2$ for all the samples. For a naive two-dimensional data set storage where each row represents one sample and each column represents one feature (i.e., one variable in BNs), we need to traverse all the rows and find four values for each row. Since $X$, $Y$, $Z_1$ and $Z_2$ are not necessarily stored next to each other, there are many random memory accesses and hence every memory access can be a cache miss.
Therefore, we instead propose to transpose the data matrix, i.e., using each row to represent one feature and each column to represent one sample, which is a cache-friendly data storage for the data. For the previous example, after the first four memory accesses of the first column, the upcoming iterations access addresses that are right next to the previously fetched values in the cache. As a result, \Name{} only has four cache misses at the beginning and the rest can be served from four cache lines.

\emph{Generating conditioning sets on-the-fly:}
In the PC-stable algorithm, processing an edge may require many CI tests, depending on the current depth $d$ and the number of adjacent nodes of its endpoints. 
In a naive implementation, we must generate all the CI tests of an edge before processing the edge. This approach is inefficient because additional memory is required to store the indices of the conditioning sets of all the CI tests. 
Given an edge $V_i - V_j$, the selection of its conditioning sets $\mathcal{S} = \{\mathcal{S}_0, \mathcal{S}_1, ..., \mathcal{S}_{\binom{p}{q}-1}\}$ can be viewed as a combination problem of choosing $q$ elements from $p = |a(V_i) \backslash \{V_j\}|$ elements at a time (cf. Algorithm~\ref{alg_pc_stable}, Line 11). \Name{} implements a combination function to generate $\mathcal{S}$ in lexicographical order~\cite{buckles1977algorithm}. Given $p$, $q$ and $r$, the combination function of \Name{} is able to directly compute the vector $\mathcal{S}_r$ without computing the whole set $\mathcal{S}$. 
With the help of the combination function, all the indices of conditioning sets of the CI tests can be computed on-the-fly and also in parallel.
Therefore, the work pool of \Name{} only contains the edges to be processed and their processing progress (i.e., $r$). No additional memory is required for storing the indices of the conditioning sets of the edges.

\subsection{Performance Analysis}

As we have discussed in Sections~\ref{sec_para} and~\ref{sec_opts}, we have optimizations for \Name{}. Here, we analyze the theoretical speedups provided by these optimizations. The optimizations to be analyzed include: (i) using CI-level parallelism with the design of the dynamic work pool; (ii) grouping the CI tests of the edges with the same endpoints; (iii) using a cache-friendly data storage. Moreover, we also use the strategy to generate conditioning sets on-the-fly. However, conditioning set generation mainly aims to reduce memory consumption, and hence we omit the speedup it provides in this section.

\subsubsection{Using CI-level parallelism with the design of the dynamic work pool}
In the depth $d$ of the graph $G$, there are $|\mathcal{E}_d|$ edges to be processed. Each edge $E_i$, with two endpoints $ep^1_i$ and $ep^2_i$, has a number of CI tests. The number of adjacent nodes of the two endpoints, denoted by $a^1_i = |adj(G, ep^1_i)|$ and $a^2_i = |adj(G, ep^2_i)|$, as well as the depth $d$ and the results of the CI tests,
determines the number of the CI tests. Specifically, each edge leads to at most $\binom{a^1_i}{d} + \binom{a^2_i}{d}$ CI tests, while if one CI test accepts the independence assumption during the processing of one edge, the process of the edge is terminated in advance (i.e., the remaining CI tests become unnecessary).

For the case of $t$ threads running in parallel, the edge-level parallelism assigns $\frac{|\mathcal{E}_d|}{t}$ edges to each thread. Ideally, the $\frac{|\mathcal{E}_d|}{t}$ edges assigned to each of the threads have the same number of CI tests to be processed. However, in practical, there is load unbalanced issue in most cases. 
For example, $\frac{|\mathcal{E}_d|}{t}$ out of the $|\mathcal{E}_d|$ edges process all the $\binom{a^1_i}{d} + \binom{a^2_i}{d}$ CI tests required for each edge $E_i$, while the other $\frac{(t-1)|\mathcal{E}_d|}{t}$ edges only process one CI test as they accept the independence assumption when handling their first CI test.  In the worst case, the $\frac{|\mathcal{E}_d|}{t}$ edges that process all the required CI tests are assigned to the same thread $p$.
In that case, the performance of the edge-level parallelism can be severely affected by this unbalanced workload, since all the threads have to wait for the completion of the slowest thread $p$. In the other words, suppose that the time for each CI test is $T_{CI}$, then the required time for the edge-level parallelism under $t$ threads is
\begin{equation} 
\label{equ_t1-1}
T_{1} = T_{CI} \sum_{i = 1}^{\frac{|\mathcal{E}_d|}{t}} (\binom{a^1_i}{d} + \binom{a^2_i}{d})
\end{equation}
However, the proposed CI-level parallelism evenly distributes all the CI tests to each thread with the help of the dynamic work pool, and hence the required time is 
\begin{equation} 
\label{equ_t1-2}
T_{2} = \frac{T_{CI}}{t} (\sum_{i = 1}^{\frac{|\mathcal{E}_d|}{t}} (\binom{a^1_i}{d} + \binom{a^2_i}{d}) + \frac{(t-1)|\mathcal{E}_d|}{t})
\end{equation}
Therefore, the speedup provided by the CI-level parallelism with the design of the dynamic work pool is
\begin{equation*} 
\label{equ_s1}
S_{CI} = \frac{T_{1}}{T_{2}}
\end{equation*}

\subsubsection{Grouping CI tests of the edges with the same endpoints} This optimization provides the speedup by reducing unnecessary CI tests. Consider the case of depth $d$ that has $|\mathcal{E}_d|$ edges to be processed, for the edge between $V_i$ and $V_j$, since edges $V_i - V_j$ and $V_j - V_i$ are viewed separately in the original PC-stable algorithm, we need to perform the CI tests considering two sets, i.e. $adj(G, V_i) \backslash \{V_j\}$ and $adj(G, V_j) \backslash \{V_i\}$. Therefore, we need to consider $2|\mathcal{E}_d|$ sets in total for the $|\mathcal{E}_d|$ edges in depth $d$. However, by grouping the CI tests of the edges $V_i - V_j$ and $V_j - V_i$, if the CI tests accept the independence hypothesis when considering the set $adj(G, V_i) \backslash \{V_j\}$, \Name{} does not consider the set $adj(G, V_j) \backslash \{V_i\}$. Suppose $\rho_d$ is the ratio of edge deletion for depth $d$. Then this optimization reduces the CI tests by $\rho_d |\mathcal{E}_d|$ unnecessary sets. That is, only $2|\mathcal{E}_d| - \rho_d |\mathcal{E}_d|$ sets need to be considered. Therefore, if we ignore the difference in the number of CI tests for different sets, the speedup brought by grouping CI tests is
\begin{equation*} 
\label{equ_s2}
S_{grouping} = \frac{2|\mathcal{E}_d|}{2|\mathcal{E}_d| - \rho_d |\mathcal{E}_d|} = \frac{2}{2 - \rho_d}
\end{equation*}

\subsubsection{Using a cache-friendly data storage}
This optimization provides the speedup by reducing the ratio of cache misses. The memory accesses of PC-stable mainly come from the accesses to the data set when computing the contingency table. For the CI test $I(X, Y| \{Z_1,..., Z_{d}\})$ in depth $d$, we need to access the values of $X$, $Y$, $Z_1$, ..., $Z_d$ of the $m$ samples in the data set, where each of the values is 4 bytes in memory. Suppose that the cache line size is $B$ bytes. Firstly we consider the access to the $\frac{B}{4}$ samples. Regarding the cache-unfriendly data storage, since $X$, $Y$, $Z_1$, ..., $Z_{d}$ are not necessarily stored next to each other, every memory access can be a cache miss. Therefore, the required time of accessing the values of $\frac{B}{4}$ samples for the cache-unfriendly data storage is
\begin{equation*} 
\label{equ_t3-1}
T_{3} = T_{DRAM} (d+2) \frac{B}{4}
\end{equation*}
where $T_{DRAM}$ represents the access time of main memory (caused by the cache misses). However, for the cache-friendly data storage, it only has $(d+2)$ cache misses for the access of the first sample, and the rest accesses of the $(\frac{B}{4}-1)$ samples can be served from the $(d+2)$ cache lines since they access addresses that are next to the previously fetched values in the cache. Therefore, the required time of accessing the values of $\frac{B}{4}$ samples for the cache-friendly data storage is
\begin{equation*} 
\label{equ_t3-2}
T_{4} = T_{DRAM} (d+2) + T_{cache} (d+2) (\frac{B}{4}-1) 
\end{equation*}
where $T_{cache}$ is the cache access time. Since $m$ is often much greater than $B$, the access time to the whole data set is a multiple of the access time to the $\frac{B}{4}$ samples. Therefore, the speedup provided by the cache-friendly data storage is
\begin{equation*} 
\label{equ_s3}
S_{cache} = \frac{T_{3}}{T_{4}}
\end{equation*}

\subsubsection{Overall speedup}
To conclude, the performance improvement of \Name{} can be computed as
\begin{equation*} 
\label{equ_s}
S = S_{CI} \cdot S_{grouping} \cdot S_{cache}
\end{equation*}
For example, let us consider the case where the number of threads $t = 4$ and the depth $d = 2$. 
Suppose that there are $|\mathcal{E}_d| = 1200$ edges at the beginning of depth 2 and 480 edges at the end, and hence the edge deletion ratio $\rho_d = 0.6$. Suppose each edge has the same number of adjacent nodes, which is the mean degree of the graph, and we assume that the mean degree is 10. Hence, every $a^1_i$ and $a^2_i$ in Equations~\eqref{equ_t1-1} and~\eqref{equ_t1-2} can be replaced by the mean degree 10. Moreover, the cache line size $B$ is often 64 bytes. The cache access time $T_{cache}$ is typically less than the access time of main memory $T_{DRAM}$ by a factor of 5 to 10, and we assume $\frac{T_{DRAM}}{T_{cache}} = 8$. 
Therefore, we can calculate the ideal speedup provided by \Name{} under these circumstances: $S_{CI} = 3.87$, $S_{grouping} = 1.43$, $S_{cache} = 5.57$, and hence the speedup $S = 30.8$. However, this theoretical analysis only provides a general speedup of \Name{}, the situation in the experiments is often more complicated than the ideal case.
For example, the values of $|\mathcal{E}_d|$, $\rho_d$, $a^1_i$ and $a^2_i$ all depend on the specific problem to be solved, and they are usually unknown beforehand.

\section{Experimental Evaluation} 
\label{sec_exp}
In this section, we conducted experiments on to evaluate the performance of our proposed techniques and compared the results to existing methods.

\subsection{Experimental Setup}
We implemented \Name{} using OpenMP in C++ for Bayesian Network structure learning and 
compared its performance to the existing methods. Specifically, we compared \Name{} with sequential implementations of the PC-stable algorithm in three different open-source packages including bnlearn~\cite{scutari2009learning}, pcalg~\cite{kalisch2012causal} and tetrad~\cite{ramsey2018tetrad}. We also compared \Name{} with the recent multi-threaded implementations in bnlearn~\cite{scutari2014bayesian} and parallel-PC~\cite{le2016fast}. Bnlearn, pcalg and parallel-PC are all R packages, while tetrad is implemented in Java. 
There are other parallel work for PC-stable, such as the work~\cite{madsen2017parallel}. However, the algorithm in~\cite{madsen2017parallel} is not open-source, and its experimental results show that its parallel implementation achieves lower speedup over its sequential implementation compared with the speedup of the parallel implementation of \Name{} over its sequential counterpart. Therefore, we did not compare \Name{} with~\cite{madsen2017parallel}. 
All the experiments were conducted on a Linux machine with two 26-core 2GHz Intel Xeon Platinum 8167M CPUs and 768GB main memory.

Data sets used in our experiments were obtained from eight benchmark BNs of different sizes listed in Table~\ref{tab_dataset}, where the last four data sets are large-scale BNs. These networks represent problems from different fields and have been widely used for comparative purposes in the literature of BN structure learning. We obtained 5,000 samples of data with no missing values from each of the networks. Besides, more data sets are obtained for the first four networks with 10,000 and 15,000 samples to test the impact of different sample sizes.
We used $G^2$ test statistic to perform the CI tests while setting the significance level $\alpha$ to 0.05 in all experiments.
The accuracy of \Name{} is exactly the same as the other PC-stable algorithm implementations, because \Name{} is an accelerated implementation of the same PC-stable algorithm. Hence, we omit reporting the results on accuracy comparison.


\begin{table}
    \centering
    \caption{BNs from which data sets used are generated.}
    \begin{tabular}{|l||c|c|c|}
        \hline
        Data set & \# of nodes & \# of edges & max \# of samples \\
        \hline
        \hline
        Alarm~\cite{beinlich1989alarm} & 37 & 46 & 15000 \\
        Insurance~\cite{binder1997adaptive} & 27 & 52 & 15000 \\
        Hepar2~\cite{onisko2003probabilistic} & 70 & 123 & 15000 \\
        Munin1~\cite{andreassen1989computer} & 186 & 273 & 15000 \\
        Diabetes~\cite{andreassen1991model} & 413 & 602 & 5000 \\
        Link~\cite{jensen1999blocking} & 724 & 1125 & 5000 \\
        Munin2~\cite{andreassen1989computer} & 1003 & 1244 & 5000 \\
        Munin3~\cite{andreassen1989computer} & 1041 & 1306 & 5000 \\
        \hline
    \end{tabular}
    \label{tab_dataset}
\end{table}

\subsection{Overall Comparison}
\label{sec_overall}

\begin{table*}
    \centering
    \caption{Execution time comparison of \Name{} with other implementations under both sequential and parallel setting. 
    Speedup of \Name{} over each compared method is also reported.}
    \begin{tabular}{|l||c|c|c|c|c|c|c||c|c|c|c|c|}
        \hline

        \multirow{3}{*}{Data set} & 
        \multicolumn{7}{c||}{Sequential implementation} &
        \multicolumn{5}{c|}{Parallel implementation} \\
        \cline{2-13}
            & 
        \multicolumn{4}{c|}{Execution time (sec)} &
        \multicolumn{3}{c||}{Speedup} &
        \multicolumn{3}{c|}{Execution time (sec)} &
        \multicolumn{2}{c|}{Speedup} \\
        \cline{2-13}
			& bnlearn & tetrad & pcalg & \Name{}
			& bnlearn & tetrad & pcalg 
			& bnlearn & parallel-PC & \Name{}
			& bnlearn & parallel-PC \\
        \hline
        \hline
        Alarm & 0.42 & 5.38 & 53.8 & 0.12 & 3.5 & 45.1 & 450 & 0.42 & 15.4 & 0.017 & 24.5 & 890 \\
        Insurance & 0.34 & 13.08 & 71.9 & 0.24 & 1.4 & 55 & 302 & 0.34 & 25.4 & 0.037 & 9.2 & 687 \\
        Hepar2 & 4.03 & 37.7 & 208.9 & 1.57 & 2.8 & 24 & 133 & 2.82 & 158 & 0.19 & 15.2 & 852 \\
        Munin1 & 111 & 770 & 2160 & 15.5 & 7.2 & 49.8 & 140 & 16.5 & 162 & 1.78 & 9.3 & 91.3 \\\cline{3-4} \cline{7-8}
        Diabetes & 
        113k  & \multicolumn{2}{c|}{$>$ 2 days} & 
        23.3k & 4.9 & \multicolumn{2}{c||}{$>$ 7.4} & 7640 & 54k & 1203 & 6.4 & 44.9 \\\cline{2-4} \cline{6-8}
        Link & \multicolumn{3}{c|}{$>$ 2 days} & 
        62.9k & \multicolumn{3}{c||}{$>$ 2.7} & 
        49.4k & $>$ 2 days & 4349 & 11.4 & $>$ 39.7 \\\cline{2-4} \cline{6-8}
        Munin2 & 27.9k & \multicolumn{2}{c|}{$>$ 2 days} & 3496 & 8.0 & \multicolumn{2}{c||}{$>$ 49.4} & 2734 & $>$ 2 days & 293 & 9.3 & $>$ 590 \\
        Munin3 & 38.7k & \multicolumn{2}{c|}{$>$ 2 days} & 8081 & 4.8 & \multicolumn{2}{c||}{$>$ 21.4} & 3621 & $>$ 2 days & 751 & 4.8 & $>$ 230 \\
        \hline
    \end{tabular}
    \label{tab_overall}
\end{table*}

In the overall evaluation of \Name{}, we compared the execution time of both sequential and parallel implementations of \Name{} with the existing implementations on the eight data sets with 5000 samples. Specifically, we compared the sequential version of \Name{} (i.e., \Name{}-seq) with the PC-stable implementations in bnlearn (i.e., bnlearn-seq)~\cite{scutari2009learning}, pcalg~\cite{kalisch2012causal} and tetrad~\cite{ramsey2018tetrad} packages; we also compared the parallel version of \Name{} (i.e., \Name{}-par) with the multi-threaded implementation in bnlearn (i.e., bnlearn-par)~\cite{scutari2014bayesian} and and parallel-PC~\cite{le2016fast}. 
The $gs$ of \Name{} was set to 1 for all the experiments here. For comparing the parallel implementations, we varied the number of OpenMP threads $t$ from 1 to 32 and chose the one with the shortest execution time. We terminated the experiment if the execution time exceeded 48 hours with no results obtained. 

The experimental results are summarized in Table~\ref{tab_overall}. As can be seen from the ``Speedup'' columns of the table, the sequential implementation of our proposed \Name{} often achieves two to three orders of magnitude speedup over tetrad and pcalg, and can be 1.4 to 7.2 times faster than the sequential version of bnlearn. 
The speedups of \Name{} are mainly due to the careful optimizations, including of grouping CI tests of the edges with the same endpoints, using a cache-friendly data storage and generating conditioning sets on-the-fly, as we discussed in Section~\ref{sec_opts}. These general optimizations can be applied to both sequential and parallel implementations.
When comparing the parallel implementations, \Name{}-par is often much faster than parallel-PC, and can run 4.8 to 24.5 times faster than bnlearn-par. It is worth noting that for some small data sets, such as \emph{Alarm} and \emph{Insurance}, bnlearn failed to get improvements by the multi-threaded techniques, and thus the same results were reported for its sequential and parallel implementations. Another observation is that \Name{} always achieves its shortest execution time when $t = 32$. Moreover, the execution time of the sequential version of \Name{} can be reduced by more than 85\% by using the parallel computing techniques. The experiment on the \emph{Link} data set is the task taking the longest time to complete. This task ran more than 2 days using the existing sequential implementations bnlearn, tetrad, pcalg and the parallel implementation parallel-PC, while the execution time is significantly reduced to about 1.2 hours using the proposed \Name{}.

To further investigate why \Name{} is faster, we used perf Linux profiler to obtain the detailed measurements for \Name{}-par, \Name{}-seq and bnlearn-par. The results on \emph{Hepars} and \emph{Munin1} are shown in Table~\ref{tab_detailed}.
We can observe that the parallel version of \Name{} increases CPU utilization and FLOPS. Moreover, compared with bnlearn, both the sequential and parallel implementations of \Name{} have fewer accesses to the L1 cache and last level (LL) cache, and significantly decreases the rate of cache misses.

\begin{table*}
    \centering
    \caption{Detailed comparison of the parallel and sequential versions of \Name{} with the parallel version of bnlearn. ``-seq'' and ``-par'' represent sequential and parallel implementation, respectively. 
    }
    \begin{tabular}{|l|c|c|c|c|c|c|}
        \hline
        
        Hepar2 & L1-cache accesses & L1-cache misses (rate) & LL-cache accesses & LL-cache misses (rate) & FLOPS & CPU utilization  \\
        \hline
        \Name{}-par & $4.5 \times 10^9$ & $7.9 \times 10^7$ (1.78\%) & $1.6 \times 10^6$ & $8.1 \times 10^4$ (5.1\%) & $1.4 \times 10^9$ & 12.7 \\
        \Name{}-seq & $4.1 \times 10^9$ & $7.2 \times 10^7$ (1.73\%) & $2.5 \times 10^5$ & $1.5 \times 10^4$ (6.0\%) & $2.3 \times 10^8$ & 1 \\
        bnlearn-par & $1.5 \times 10^{10}$ & $4.7 \times 10^8$ (3.17\%) & $4.2 \times 10^7$ & $1.7 \times 10^7$ (39.9\%) & $7.0 \times 10^7$ & 3.7 \\
        \hline
        \hline
        Munin1 & L1-cache accesses & L1-cache misses (rate) & LL-cache accesses & LL-cache misses (rate) & FLOPS & CPU utilization  \\
        \hline
        \Name{}-par & $3.8 \times 10^{10}$ & $8.7 \times 10^8$ (2.28\%) & $8.9 \times 10^6$ & $1.8 \times 10^5$ (2.03\%) & $2.3 \times 10^9$ & 13.2  \\
        \Name{}-seq & $3.8 \times 10^{10}$ & $8.8 \times 10^8$ (2.28\%) & $8.6 \times 10^6$ & $9.3 \times 10^4$ (1.08\%) & $2.7 \times 10^8$ & 1 \\
        bnlearn-par & $1.0 \times 10^{11}$ & $3.0 \times 10^9$ (2.92\%) & $1.7 \times 10^8$ & $8.2 \times 10^7$ (47.1\%) & $2.4 \times 10^8$ & 8.7  \\
        \hline
    \end{tabular}
    \label{tab_detailed}
\end{table*}

\subsection{Studies on Different Granularities}

Compared with the schemes of the edge-level and sample-level parallelism, the CI-level parallelism employed in \Name{} solves the efficiency issues of load unbalancing, atomic operations and large parallel overhead (cf. Sections~\ref{sec_para} and~\ref{sec_ours}). 
To investigate the performance of different parallelism granularities, we implemented another two parallel versions using the schemes of edge-level and sample-level parallelism, and compared it with \Name{} that employs the CI-level parallelism. All these parallel versions are based on the optimized sequential version of \Name{}.

\begin{figure}
	\centering
	\subfloat[Alarm]
	{\includegraphics[width=1.7in]{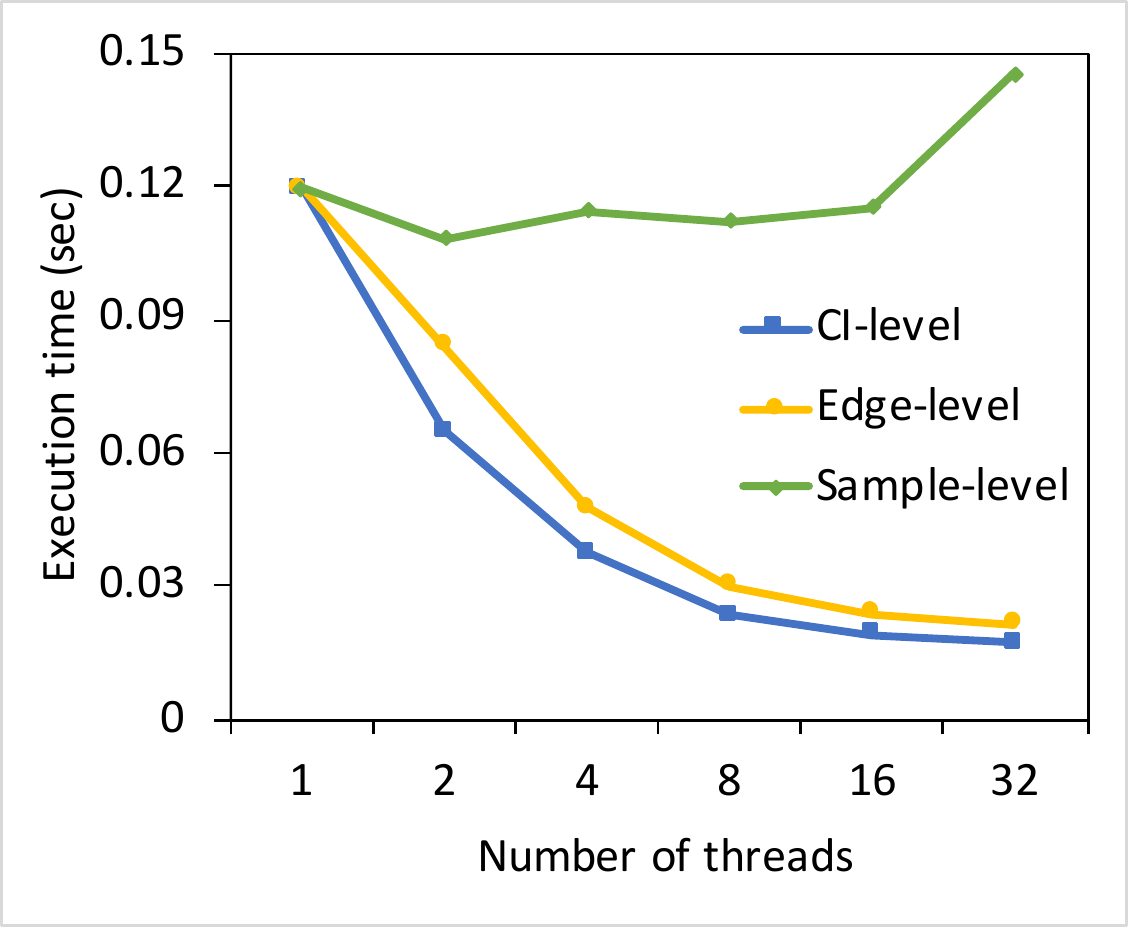}
		\label{fig_alarm_para}} 
	\subfloat[Insurance]
	{\includegraphics[width=1.7in]{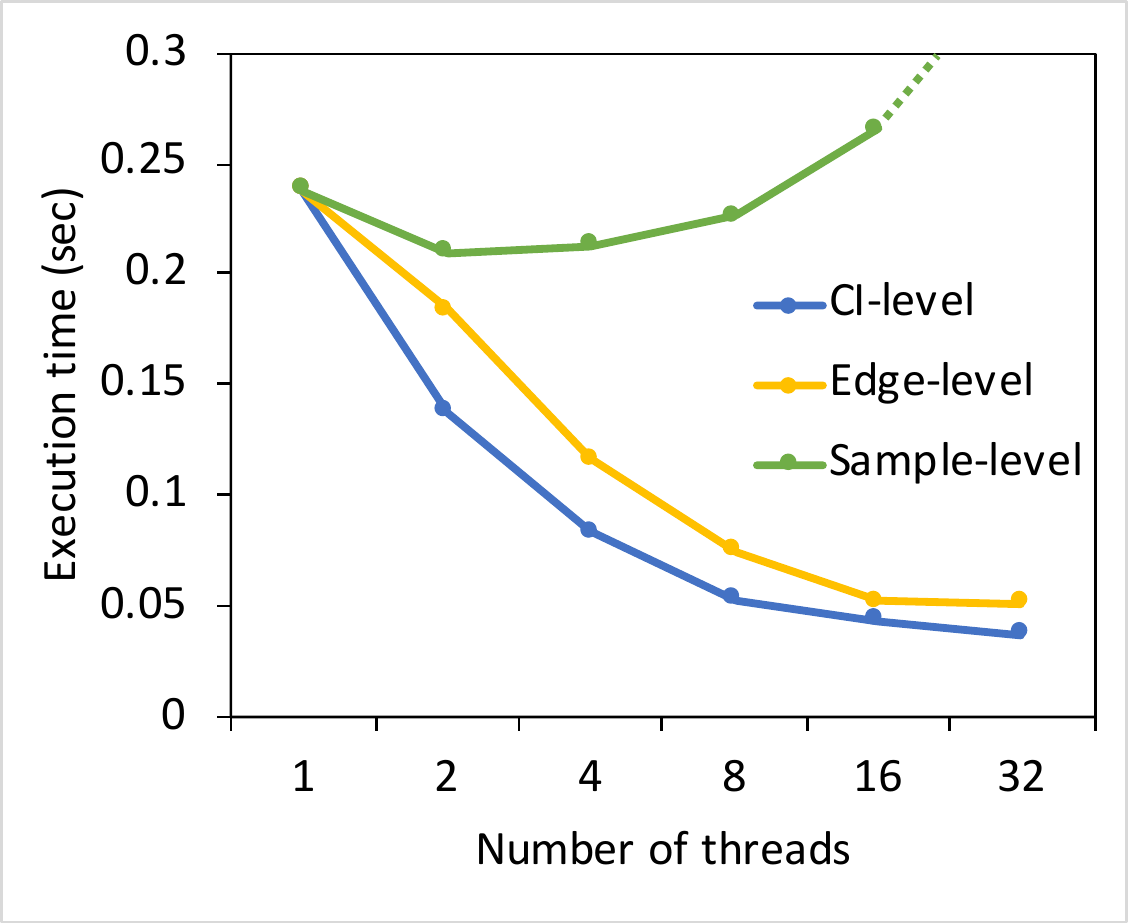} 
		\label{fig_insurance_para}} 
	\hfill
	
	\subfloat[Hepar2]
	{\includegraphics[width=1.7in]{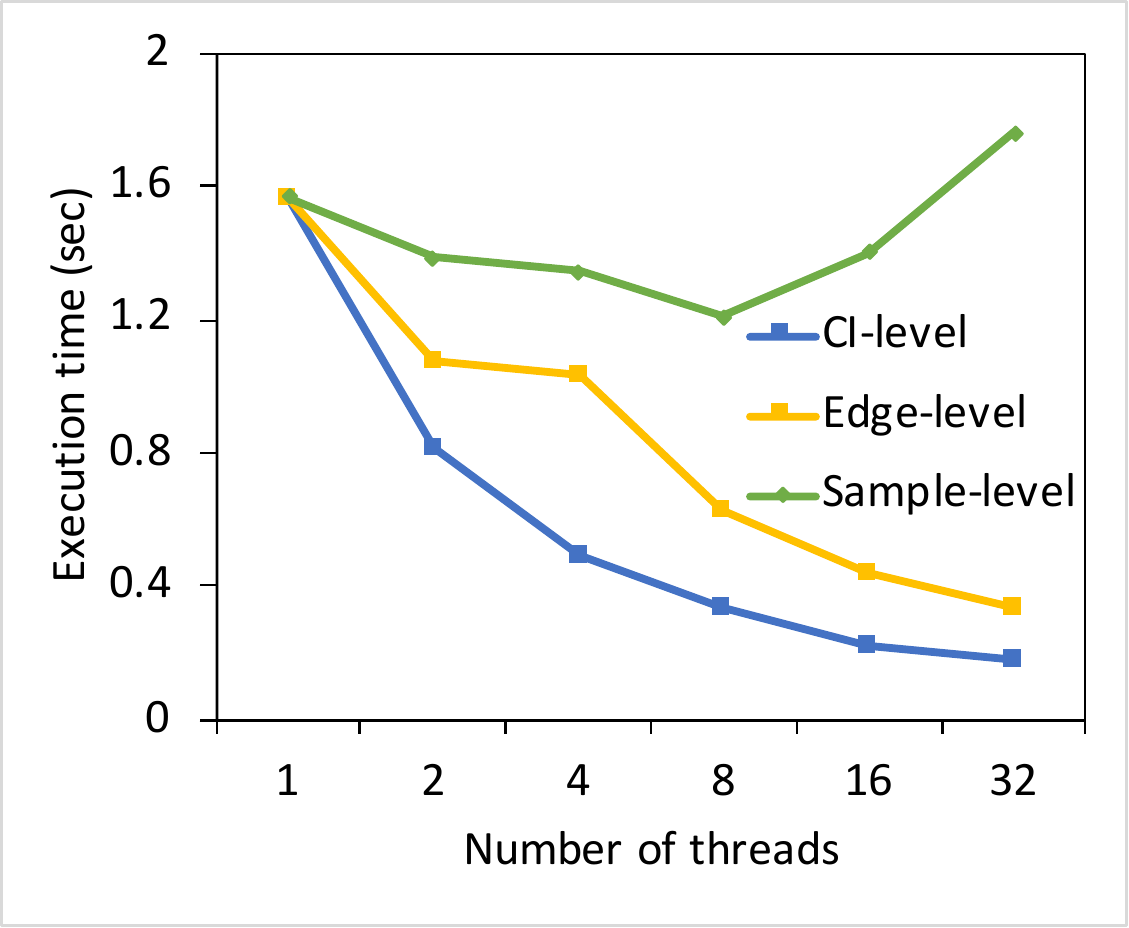} 
		\label{fig_hepar2_para}} 
	\subfloat[Munin1]
	{\includegraphics[width=1.7in]{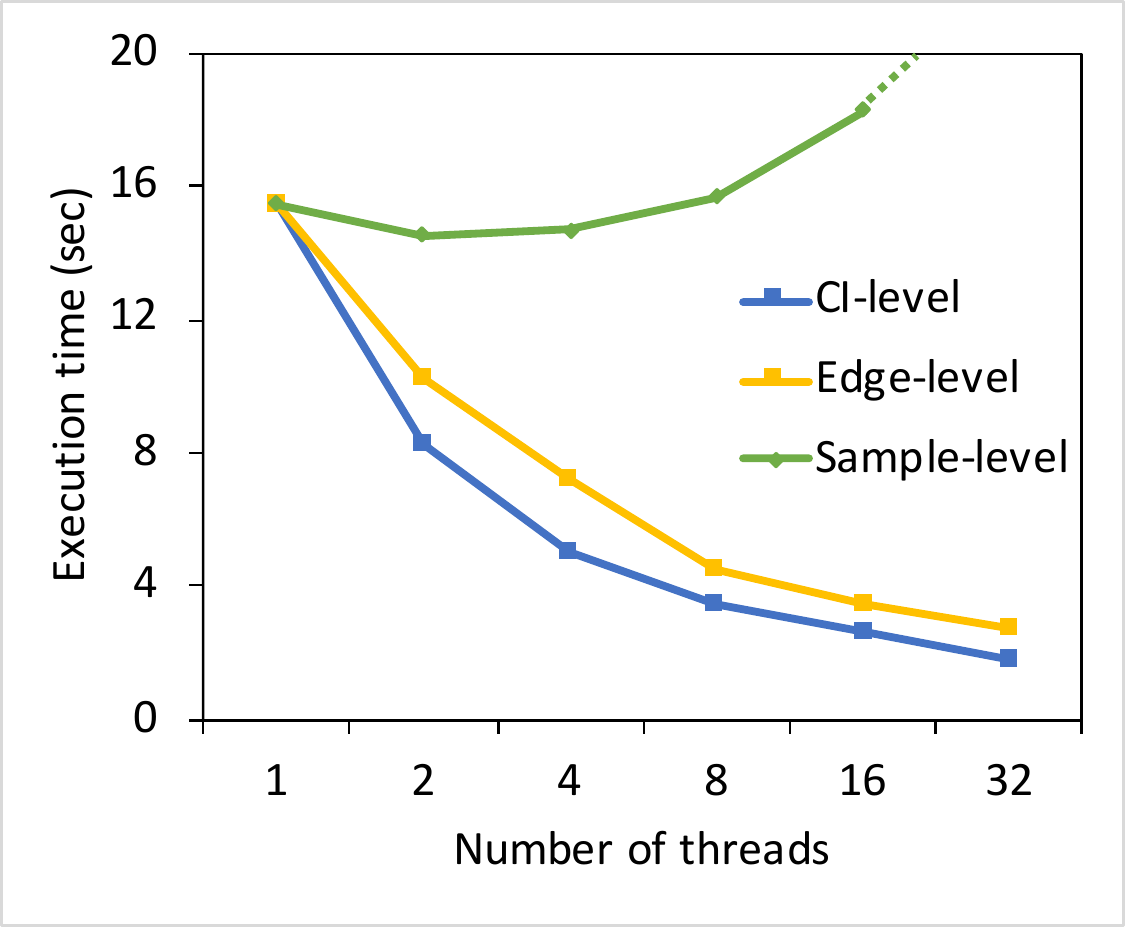} 
		\label{fig_munin1_para}} 
		
	\subfloat[Diabetes]
	{\includegraphics[width=1.7in]{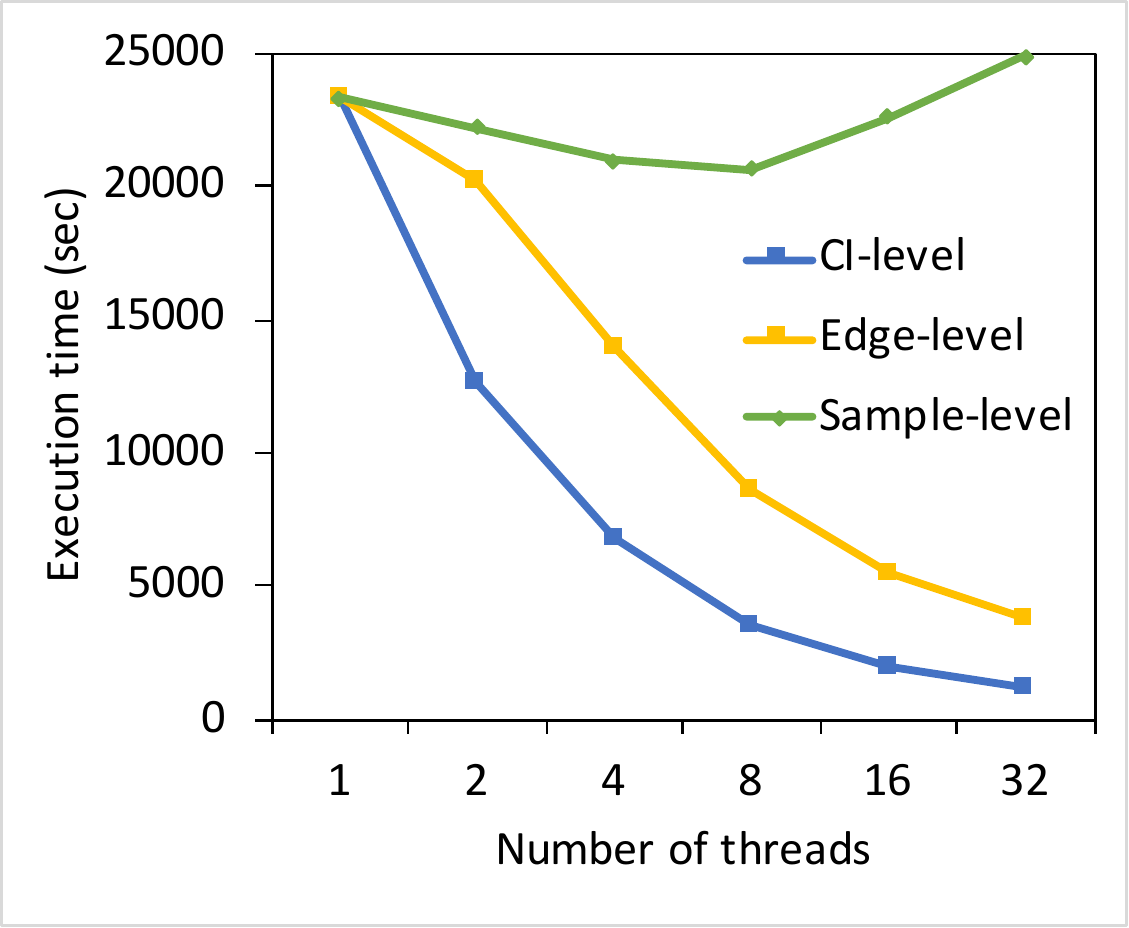} 
		\label{fig_diabetes_para}} 
	\subfloat[Link]
	{\includegraphics[width=1.7in]{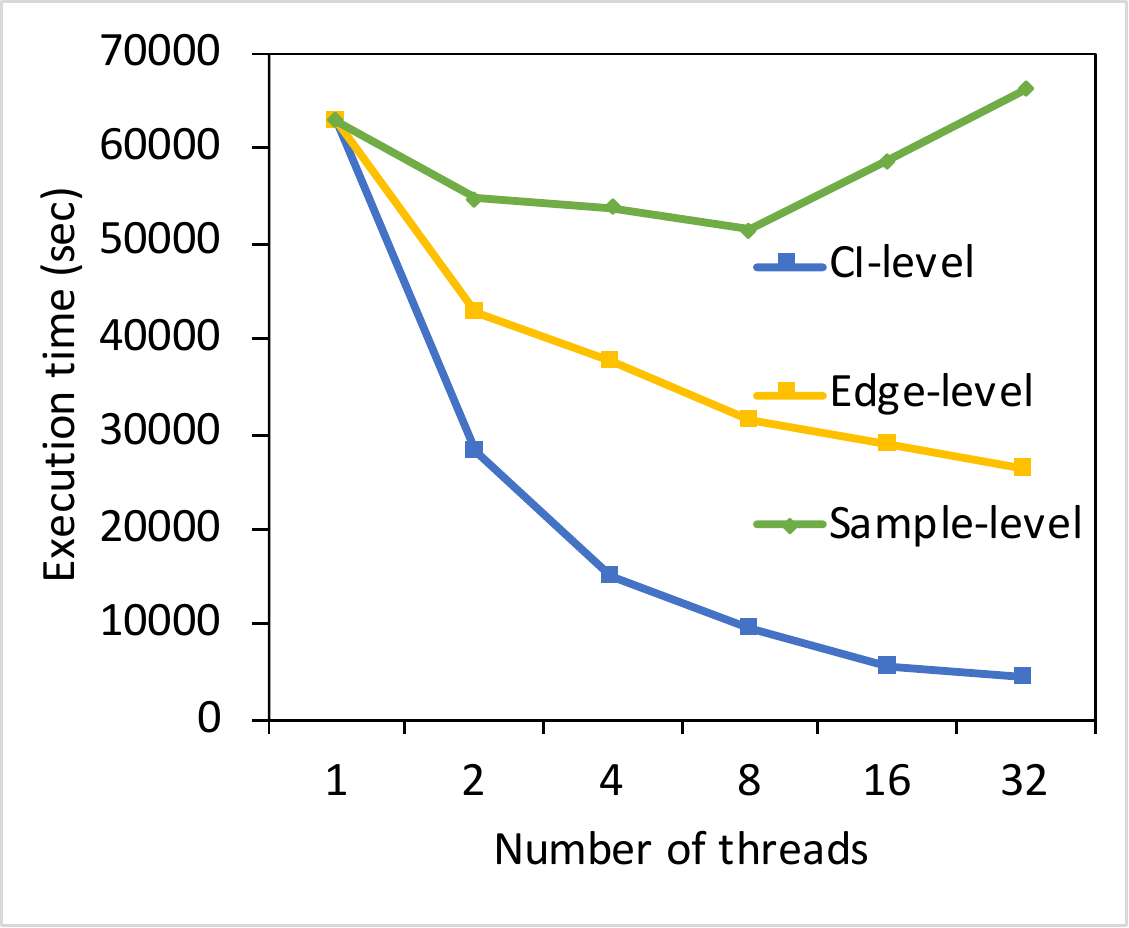} 
		\label{fig_link_para}} 
	
	\caption{Execution time of parallel implementations using three different granularities of parallelism: CI-level parallelism, edge-level parallelism and sample-level parallelism.}  
	\label{fig_para} 
\end{figure}

Figure~\ref{fig_para} illustrates the execution time of the three schemes using different granularities of parallelism with different number of threads.
We observe that the CI-level parallelism always leads to the shortest execution time under different number of threads, indicating the effectiveness of the optimizations used in the CI-level parallelism. 
Overall, the sample-level parallelism is the worst due to the efficiency issues of many expensive atomic operations and large parallelization overhead.
Moreover, the execution time of the edge-level parallelism can be reduced by more than 20\% using the CI-level parallelism which solves its load unbalancing issue. Learning larger-scale BNs may encounter more issues of load unbalancing, and \Name{} can take more advantage of the load balancing optimization in the CI-level parallelism. On \emph{Diabetes} and \emph{Link}, the improvement is over 3 times.

\subsection{Sensitivity Studies}

To better understand \Name{}, we performed sensitivity studies on three key parameters: sample size, group size and network size. A series of experiments were carried out by changing the sample size and group size. Experiments were also conducted on data sets with different network sizes.

\begin{figure}
	\centering
	\subfloat[Alarm]
	{\includegraphics[width=1.7in]{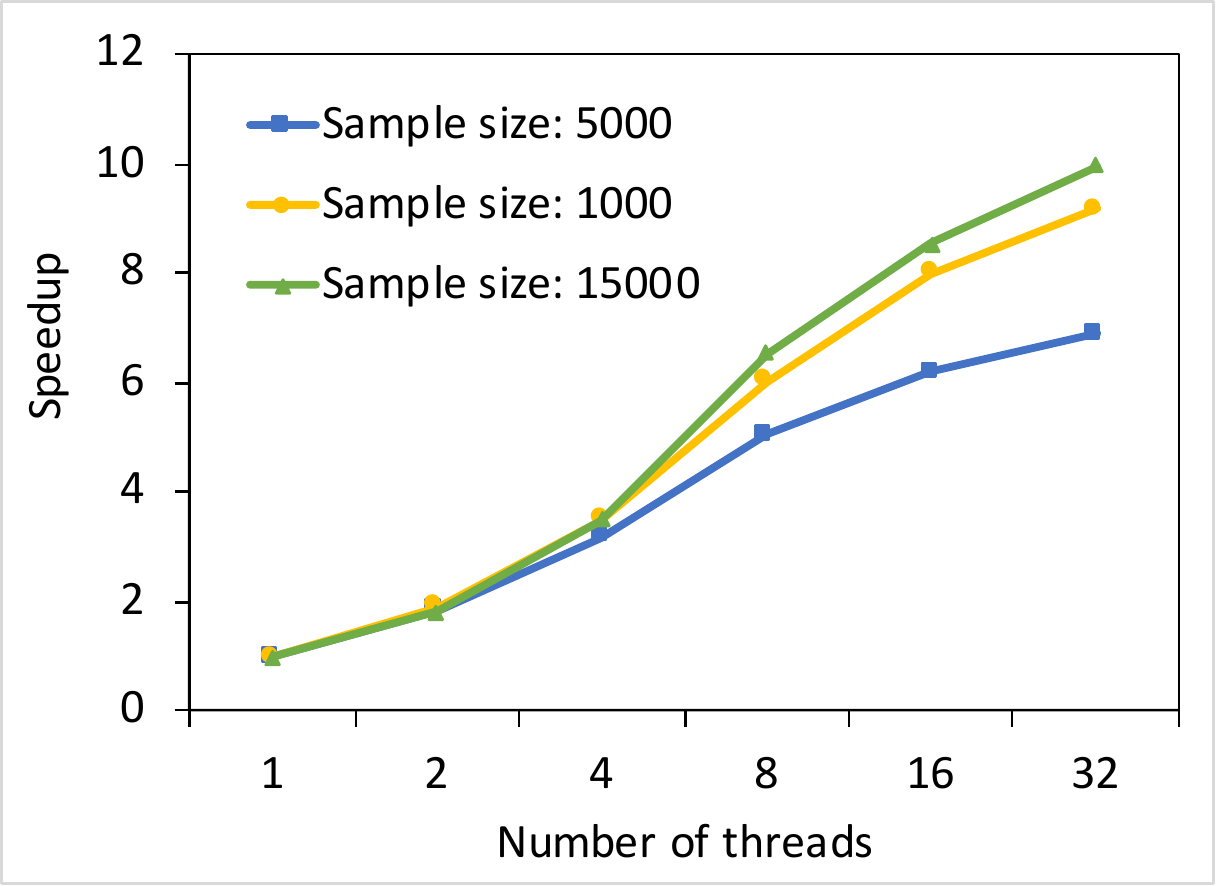}
		\label{fig_alarm_sample_size}} 
	\subfloat[Insurance]
	{\includegraphics[width=1.7in]{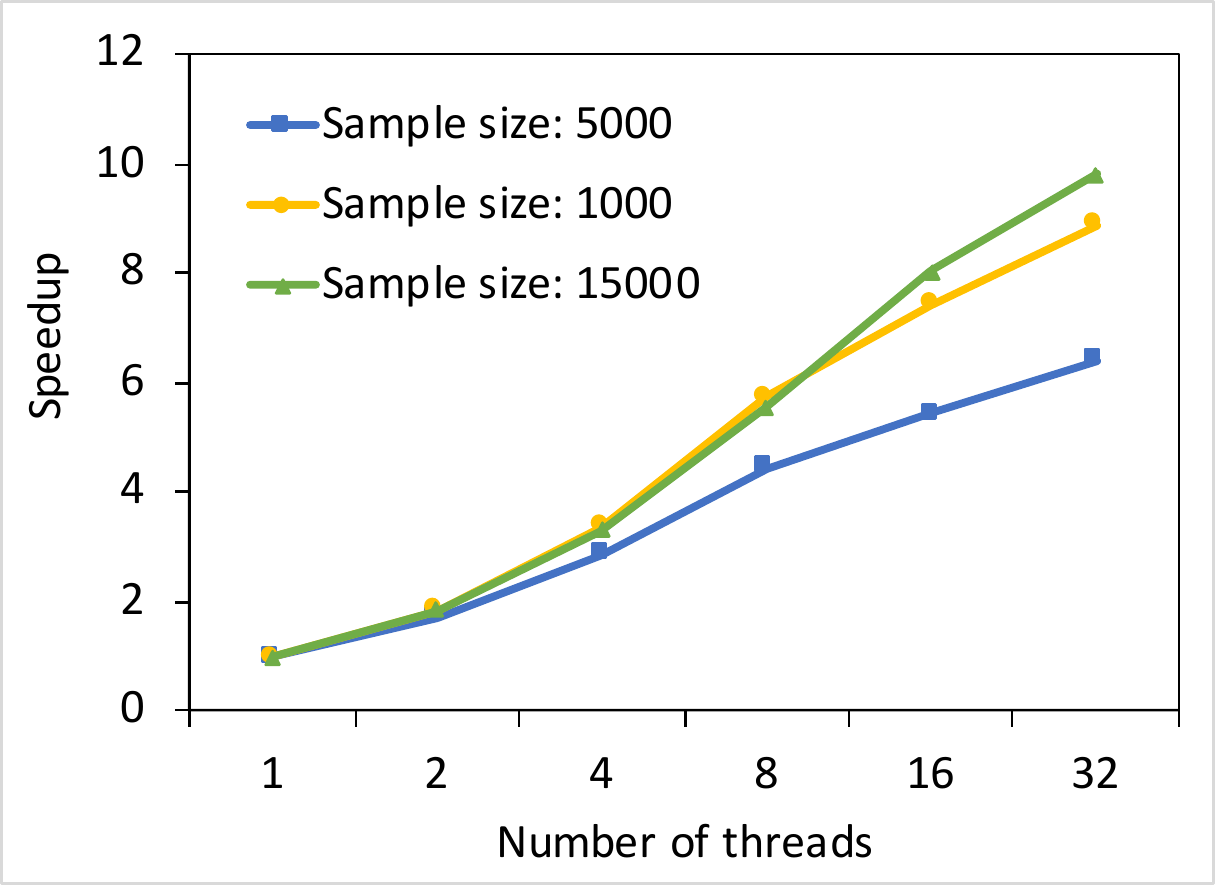} 
		\label{fig_insurance_sample_size}} 
	\hfill
	
	\subfloat[Hepar2]
	{\includegraphics[width=1.7in]{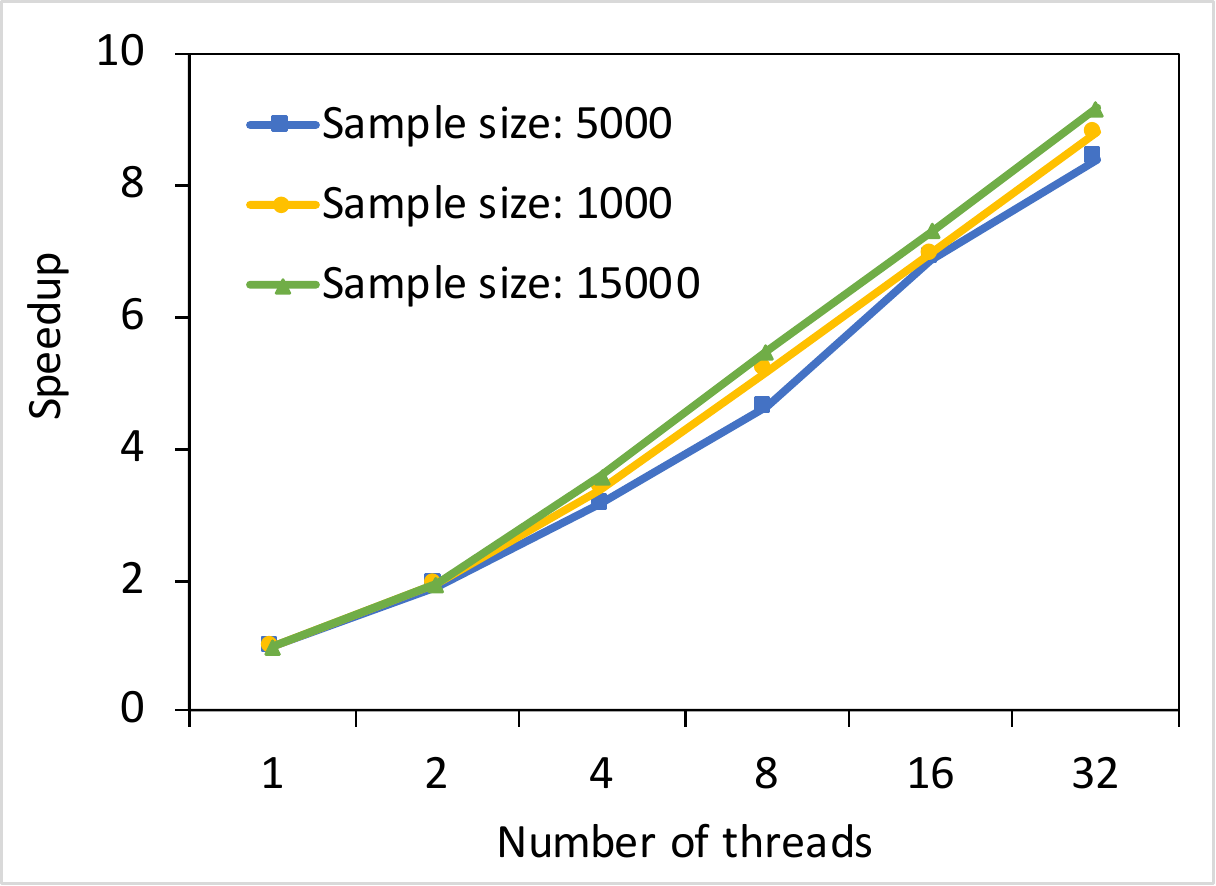} 
		\label{fig_hepar2_sample_size}} 
	\subfloat[Munin1]
	{\includegraphics[width=1.7in]{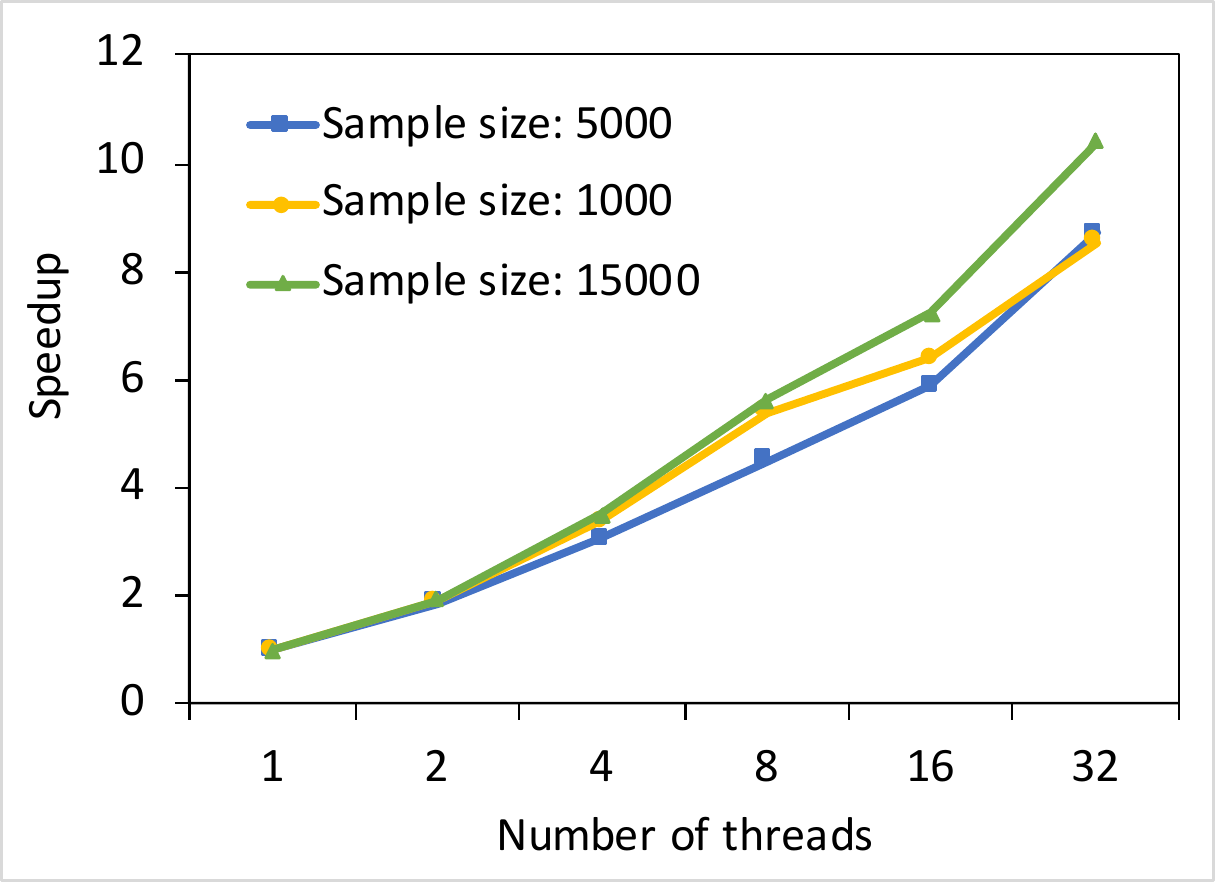} 
		\label{fig_munin1_sample_size}} 
	
	\caption{Studies on speedup of \Name{}-par over \Name{}-seq under different sample sizes.}  
	\label{fig_sample_size} 
\end{figure}

\emph{Varying sample size:}
We conducted experiments on \emph{Alarm}, \emph{Insurance}, \emph{Hepar2} and \emph{Munin1} networks to investigate the scalability of the proposed \Name{} to the sample size. We used different data sets of 5,000, 10,000 and 15,000 samples, and compared the execution time for the different sample sizes. Figure~\ref{fig_sample_size} shows the
speedups of the parallel implementation of \Name{} over the sequential implementation of \Name{} under different sample sizes.
We can observe a smooth improvement in speedups for all the sample sizes, indicating good scalability of the proposed techniques to the sample size. A large sample size often gets a slightly higher speedup because in this case, each CI test has a larger amount of workload which can better amortize the parallel overhead.

\begin{figure}
	\centering
	\subfloat[Alarm]
	{\includegraphics[width=1.7in]{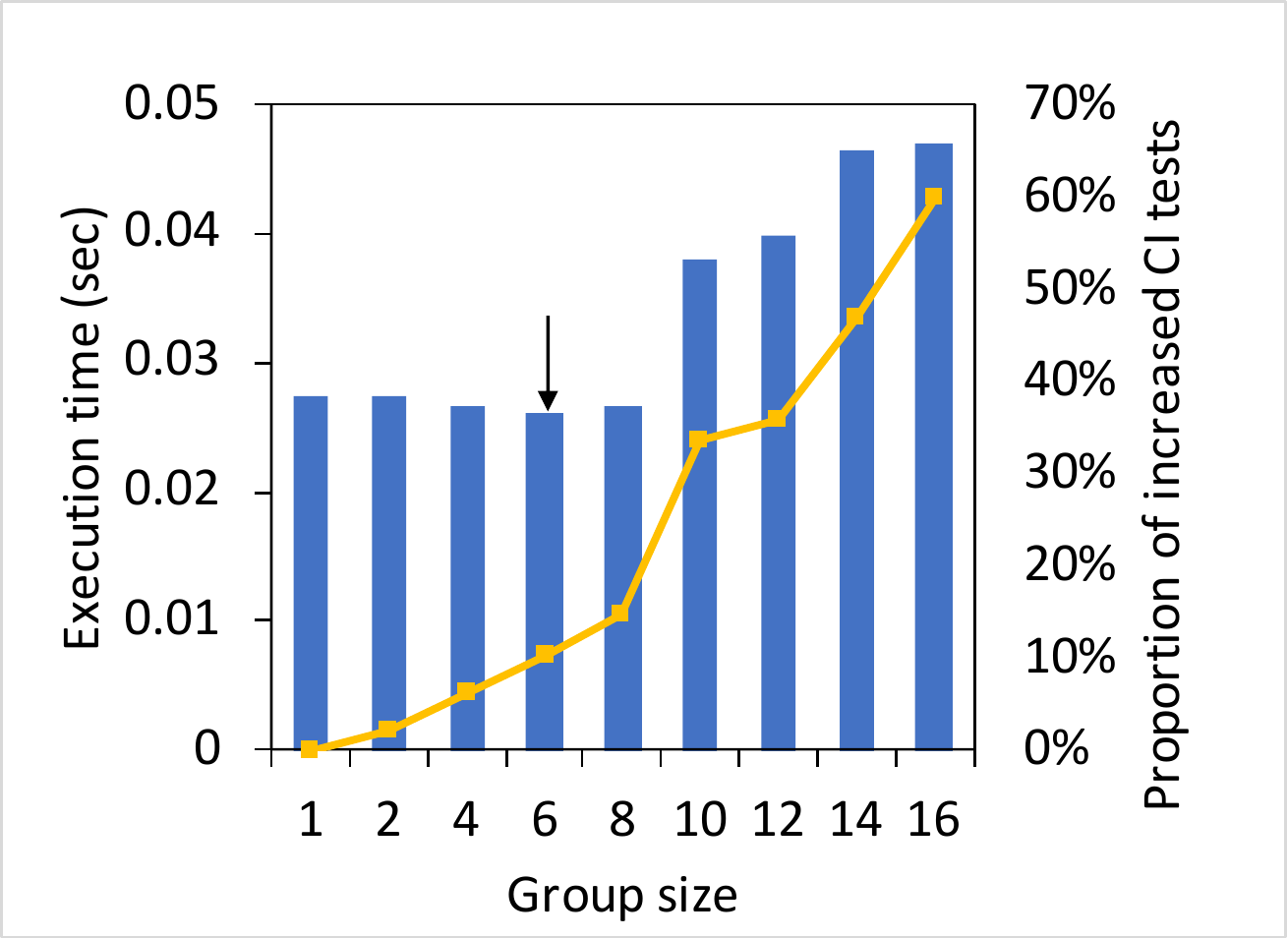}
		\label{fig_alarm_group_size}} 
	\subfloat[Insurance]
	{\includegraphics[width=1.7in]{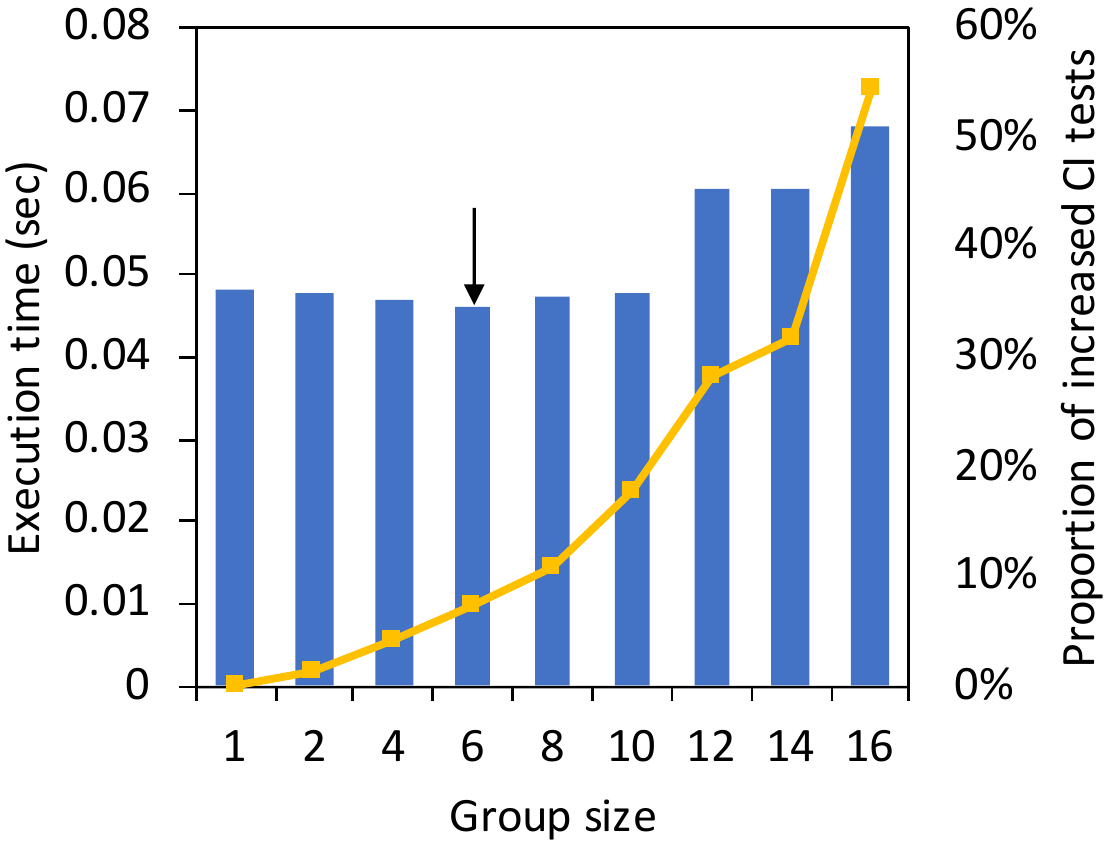} 
		\label{fig_insurance_group_size}} 
	\hfill
	
	\subfloat[Hepar2]
	{\includegraphics[width=1.7in]{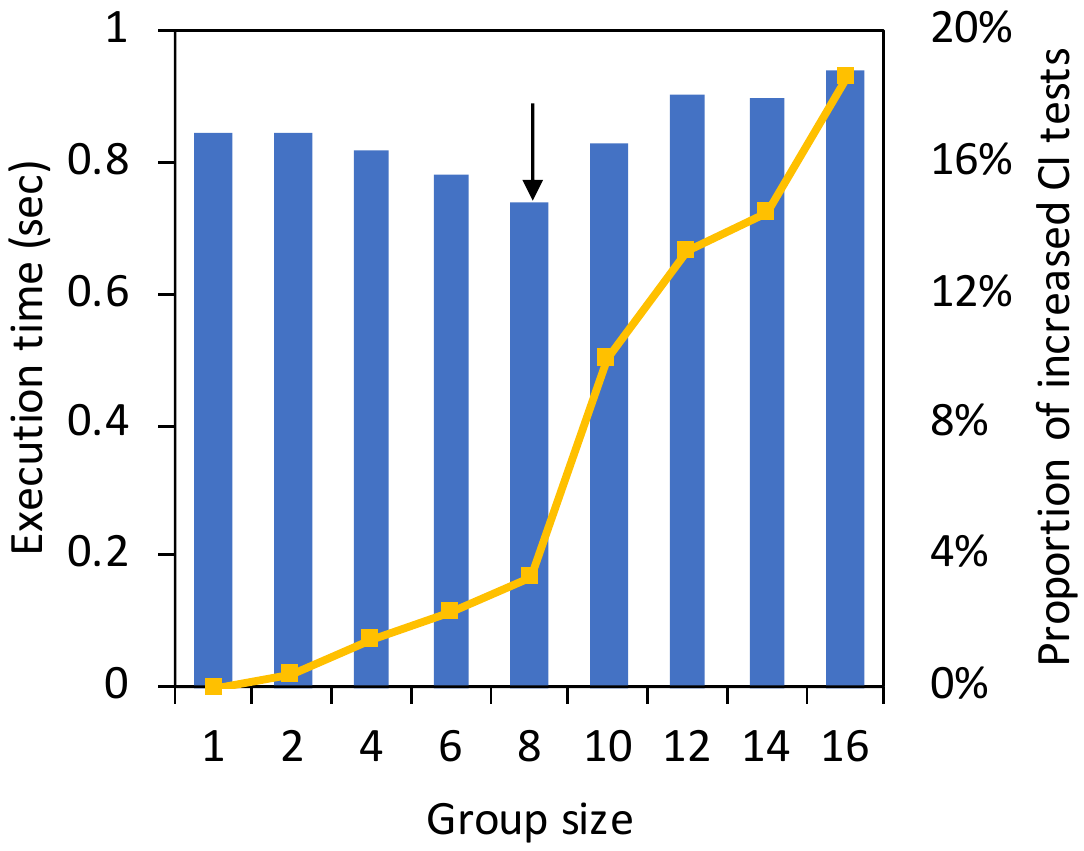} 
		\label{fig_hepar2_group_size}} 
	\subfloat[Munin1]
	{\includegraphics[width=1.7in]{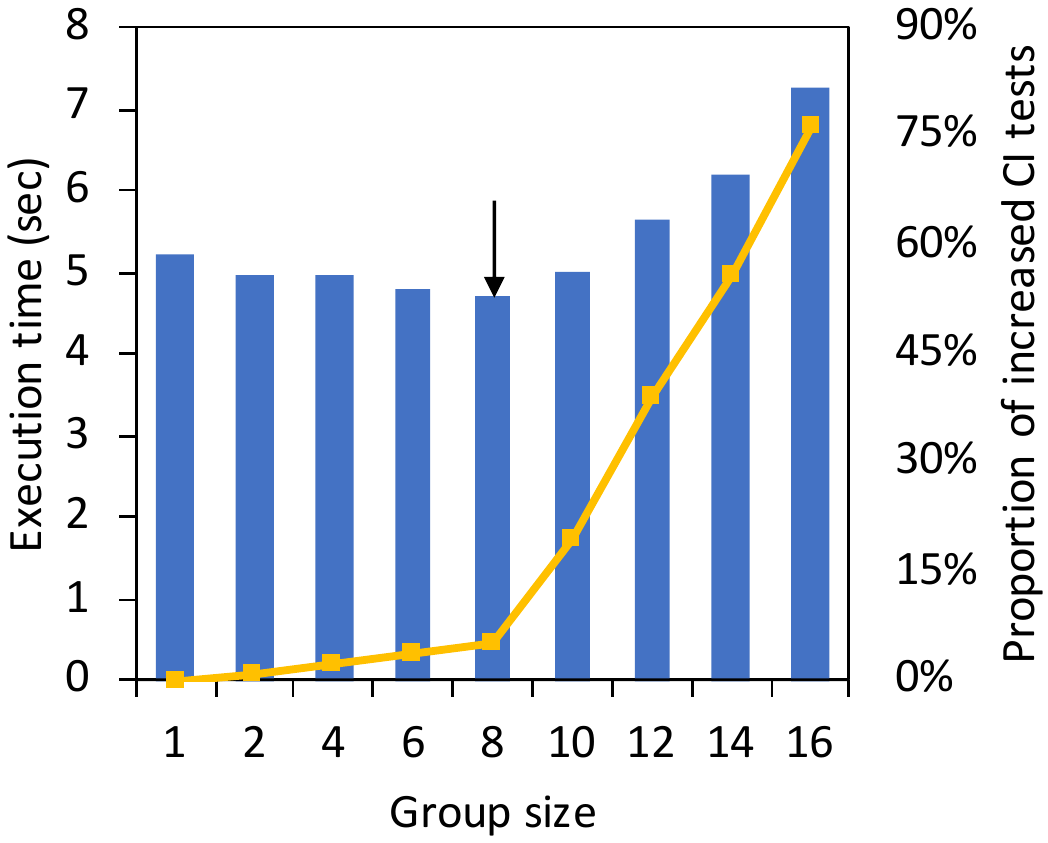} 
		\label{fig_munin1_group_size}} 
	
	\caption{Effect of group sizes on execution times and the number of increased CI tests.}  
	\label{fig_group_size} 
\end{figure}

\emph{Varying group size:}
We examined the effect of $gs$ of \Name{} on the number of CI tests and execution time. We tested on \emph{Alarm}, \emph{Insurance}, \emph{Hepar2} and \emph{Munin1} networks with 10,000 samples, which are also used in the previous experiments. The results are shown in Figure~\ref{fig_group_size}. The bars in the figure illustrate the execution time under different group sizes for the four problems, and the lines illustrate the proportion of the CI tests increased by the group size compared to the case of $gs = 1$. 
We can observe that the number of CI tests increases with the increase in group size. Therefore, although a larger group size can reduce more memory accesses, the final execution time may increase for the larger group sizes. The $gs$ is a trade-off between the number of CI tests and memory accesses. Our observation is that the proportion of the increased CI tests is not too high (e.g., less than 10\%) when the group size is no more than 8, while the number of CI tests often increases more rapidly when the group size is greater than 8. For example, the proportion of the increased CI tests is about 5\% for \emph{Munin1} when $gs = 8$, while the proportion increases to about 20\% when $gs = 10$. As a result, the shortest execution time is often achieved when $gs \leq 8$ and it depends on the specific problem. The downward arrows in Figure~\ref{fig_group_size} mark the $gs$ that achieves the shortest execution time. We can observe that \emph{Alarm} and \emph{Insurance} achieve the shortest execution time when $gs = 6$, and \emph{Hepar2} and \emph{Munin1} achieve the shortest execution time when $gs = 8$. 
Note that the experimental results of \Name{} in Section~\ref{sec_overall} are with $gs = 1$, and thus the execution time can be further reduced by about 10\% with careful parameter tuning.

\begin{figure}
	\centering
	\includegraphics[width=0.38\textwidth]{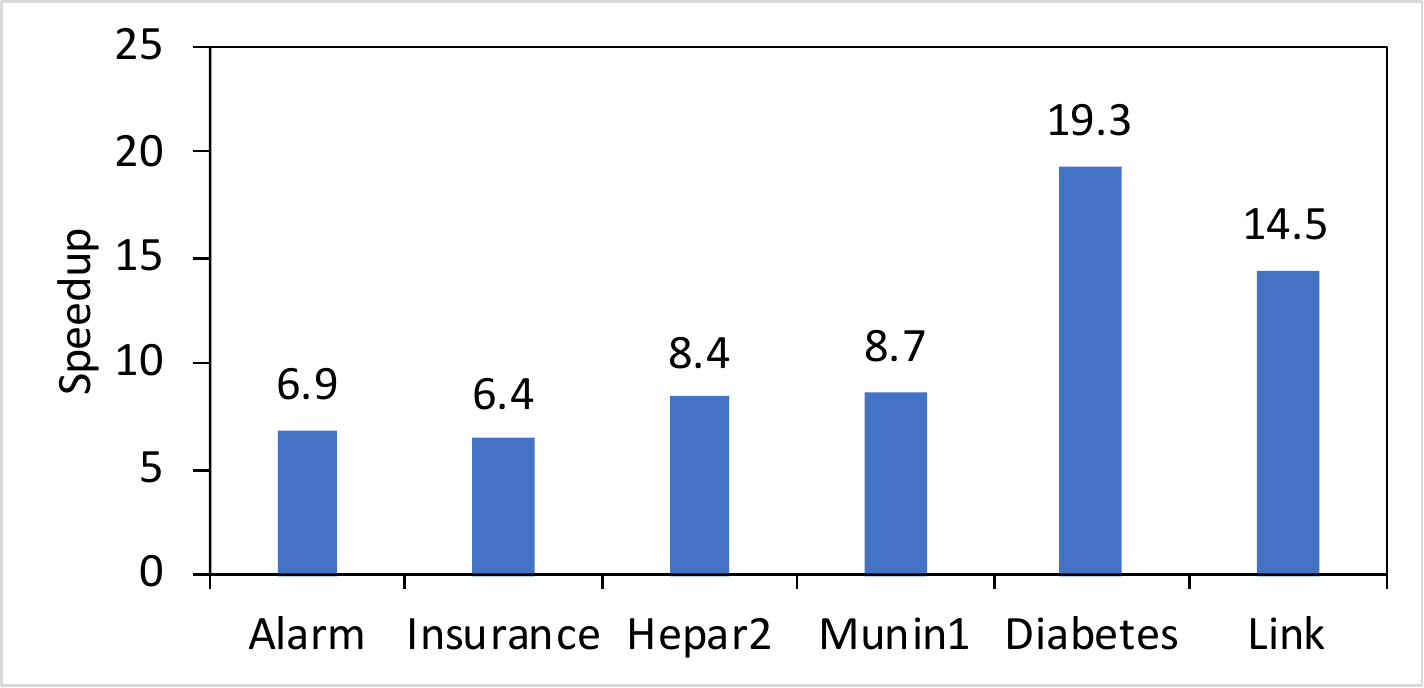}
	\caption{Studies on speedup of \Name{}-par over \Name{}-seq under different network sizes.}
	\label{fig_speedup}
\end{figure}

\emph{Different network sizes:}
Figure~\ref{fig_speedup} shows the speedups of the parallel implementation of \Name{} over its sequential implementation on the data sets with 5000 samples.
The six BNs tested in the experiments are of different network sizes as shown in Table~\ref{tab_dataset}. 
We can observe that \Name{} can achieve high speedups for large-scale networks, indicating good scalability of the proposed techniques to large networks. For example, \Name{} achieves 19.3 times speedup on the \emph{Diabetes} network, which contains 413 nodes and 602 edges. For small-scale networks, the speedups of parallel implementation is relatively smaller, because they already require short execution time for structure learning (e.g., less than 1 second for \emph{Alarm} and \emph{Insurance}) and the parallelization overhead of these small-scale networks accounts for a large proportion.
Therefore, with the help of the proposed general optimizations discussed in Section~\ref{sec_opts},
our sequential implementation is sufficient for such small-scale networks with a relatively small number of nodes and edges.

\section{Conclusion}
\label{sec_conclusion}

In this paper, we have proposed a parallel PC-stable algorithm namely \Name{} for learning Bayesian Network (BN) structure. The challenges of developing a fast solution for BN structure learning include addressing load unbalancing issues, reducing atomic operations and amortizing parallel overhead. To tackle these challenges, \Name{} exploits the CI-level parallelism, which avoids the expensive atomic operations and has suitable amount of workloads to amortize the parallel overhead. A dynamic work pool is designed to monitor the processing progress of edges and to schedule the workloads of threads, so as to balance the workload among threads. \Name{} also leverages a series of techniques to reduce the unnecessary CI tests and improve the memory efficiency. We have conducted extensive experiments to test the effectiveness of \Name{}. Experimental results have shown that the sequential version of \Name{} is up to 50 times faster than the existing solutions. When compared with the parallel counterpart, the parallel version of \Name{} is 9.2 to 24.5 times faster. Furthermore, \Name{} has demonstrated good scalability to network size and sample size. 


\section*{Acknowledgment}
Professor Ajmal Mian is the recipient of an Australian Research Council Future Fellowship Award (project number FT210100268) funded by the Australian Government. This research is also supported by Oracle for Research, Australia.

\bibliography{myref}
\bibliographystyle{IEEEtran}

\end{document}